\documentclass[10pt,twocolumn,letterpaper]{article}

\usepackage{cvpr}
\usepackage{times}
\usepackage{epsfig}
\usepackage{graphicx}
\usepackage{amsmath}
\usepackage{amssymb}
\usepackage[inline]{enumitem}
\usepackage{makecell}  
\usepackage{multirow}
\usepackage{gensymb}
\usepackage{soul}
\usepackage[dvipsnames]{xcolor}
\usepackage{calc}
\usepackage{booktabs}

\usepackage[pagebackref=true,breaklinks=true,letterpaper=true,colorlinks,bookmarks=false]{hyperref}

\newcommand\hlf[1]{\textbf{\textcolor{red}{#1}}} 
\newcommand\hls[1]{\textbf{\textcolor{blue}{#1}}}  

\newcommand{\0}{\phantom{0}}
\newcommand{\PAR}[1]{\vskip4pt \noindent{\bf #1~}}

\let\oldenumerate\enumerate
\renewcommand{\enumerate}{
\oldenumerate
\setlength{\itemsep}{1.2pt}
\setlength{\parskip}{0pt}
\setlength{\parsep}{0pt}
}

\def\eg{\emph{e.g.\ }}
\def\ie{\emph{i.e.\ }}

\cvprfinalcopy 

\ifcvprfinal\fi
\begin{document}

\title{From Coarse to Fine: Robust Hierarchical Localization at Large Scale}

\author{Paul-Edouard Sarlin$^1$ \quad
Cesar Cadena$^1$ \quad
Roland Siegwart$^1$ \quad
Marcin Dymczyk$^{1,2}$ \\
$^1$Autonomous Systems Lab, ETH Z\"{u}rich \quad 
$^2$Sevensense Robotics AG
}

\maketitle
\thispagestyle{empty}

\begin{abstract}
Robust and accurate visual localization is a fundamental capability for numerous applications, such as autonomous driving, mobile robotics, or augmented reality.
It remains, however, a challenging task, particularly for large-scale environments and in presence of significant appearance changes.
State-of-the-art methods not only struggle with such scenarios, but are often too resource intensive for certain real-time applications.
In this paper we propose HF-Net, a hierarchical localization approach based on a monolithic CNN that simultaneously predicts local features and global descriptors for accurate 6\nobreakdash-DoF localization.
We exploit the coarse-to-fine localization paradigm: we first perform a global retrieval to obtain location hypotheses and only later match local features within those candidate places.
This hierarchical approach incurs significant runtime savings and makes our system suitable for real-time operation.
By leveraging learned descriptors, our method achieves remarkable localization robustness across large variations of appearance and sets a new state-of-the-art on two challenging benchmarks for large-scale localization.\footnote{Code available at \url{https://github.com/ethz-asl/hf_net}}
\end{abstract}

\vspace{-5mm}
\section{Introduction}

The precise 6-Degree-of-Freedom (DoF) localization of a camera within an existing 3D model is one of the core computer vision capabilities that unlocks a number of recent applications. These include autonomous driving in GPS-denied environments~\cite{burki2019vizard, mcmanus2014shady, milford2012seqslam, barsan2018learning} and consumer devices with augmented reality features~\cite{middelberg, klein2007parallel}, where a~centimeter-accurate 6-DoF pose is crucial to guarantee reliable and safe operation and fully immersive experiences, respectively. More broadly, visual localization is a key component in computer vision tasks such as Structure-from-Motion (SfM) or SLAM. This growing range of applications of visual localization calls for reliable operation both indoors and outdoors, irrespective of the weather, illumination, or seasonal changes.

Robustness to such large variations is therefore critical, along with limited computational resources. Maintaining a model that allows accurate localization in multiple conditions, while remaining compact, is thus of utmost importance. In this work, we investigate whether it is actually possible to robustly localize in large-scale changing environments with constrained resources of mobile devices. More specifically, we aim at estimating the 6-DoF pose of a query image w.r.t. a given 3D model with the highest possible accuracy.

\begin{figure}[t]
   \centering
   \includegraphics[width=1.0\linewidth]{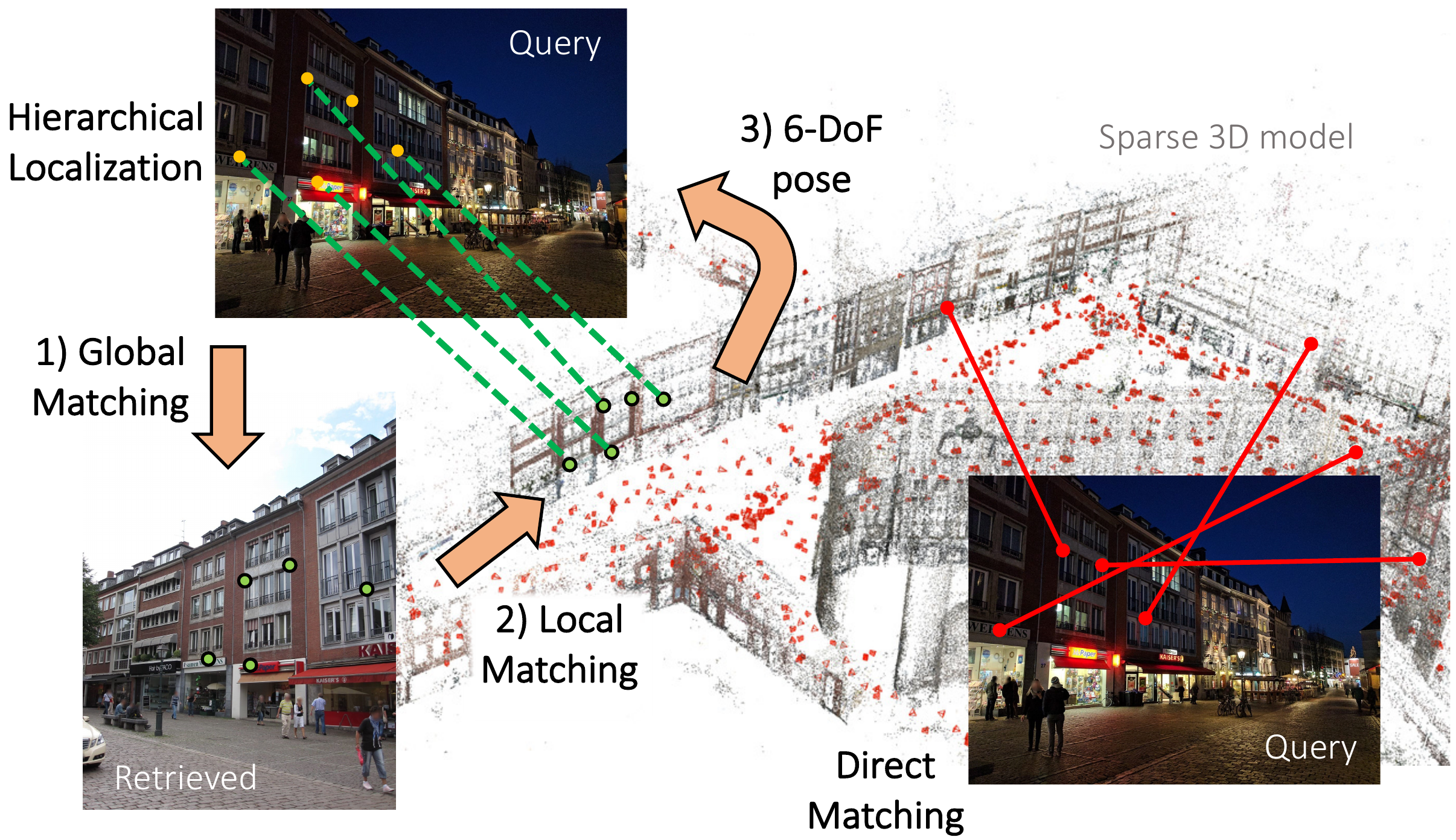}%
   \caption{\textbf{Hierarchical localization.} A global search first retrieves candidate images, which are subsequently matched using powerful local features to estimate an accurate 6-DoF pose. This two-step process is both efficient and robust in challenging situations.}
\label{fig:teaser}
\end{figure}

Current leading approaches mostly rely on estimating correspondences between 2D keypoints in the query and 3D points in a sparse model using local descriptors. This direct matching is either robust but intractable on mobile~\cite{city-scale-loc, toft2018semantic, active-search}, or optimized for efficiency but fragile~\cite{get-out-of-my-lab}. In both cases, the robustness of classical localization methods is limited by the poor invariance of hand-crafted local features~\cite{calonder2010brief, lowe2004distinctive}. Recent features emerging from convolutional neural networks~(CNN) exhibit unrivalled robustness at a low compute cost~\cite{superpoint, Fathy_2018_ECCV, mishchuk2017working}. They have been, however, only recently~\cite{inloc} applied to the visual localization problem, and only in a dense, expensive manner. Learned sparse descriptors~\cite{superpoint, ono2018lf} promise large benefits that remain yet unexplored in localization.

Alternative localization approaches based on image retrieval have recently shown promising results in terms of robustness and efficiency, but are not competitive in terms of accuracy. The benefits of an intermediate retrieval step have been demonstrated earlier~\cite{corl}, but fall short of reaching the scalability required by city-scale localization.

In this paper, we propose to leverage recent advances in learned features to bridge the gap between robustness and efficiency in the hierarchical localization paradigm. Similar to how humans localize, we employ a natural coarse-to-fine pose estimation process which leverages both global descriptors and local features, and scales well with large environments (Figure~\ref{fig:teaser}). We show that learned descriptors enable unrivaled robustness in challenging conditions, while learned keypoints improve the efficiency in terms of compute and memory thanks to their higher repeatability.
To further improve the efficiency of this approach, we propose a Hierarchical Feature Network~(HF-Net), a CNN that jointly estimates local and global features, and thus maximizes the sharing of computations. We show how such a compressed model can be trained in a flexible way using multitask distillation. By distilling multiple state-of-the-art predictors jointly into a single model, we obtain an incomparably fast, yet robust and accurate, localization. Such heterogenous distillation is applicable beyond visual localization to tasks that require both multimodal expensive predictions and computational efficiency. 
Overall, our contributions are as follows:
\begin{itemize}[label={--}]
    \item We set a new state-of-the-art in several public benchmarks for large-scale localization with an outstanding robustness in particularly challenging conditions;
    \item We introduce HF-Net, a monolithic neural network which efficiently predicts hierarchical features for a~fast and robust localization;
    \item We demonstrate the practical usefulness and effectiveness of multitask distillation to achieve runtime goals with heterogeneous predictors.
\end{itemize}
\section{Related Work}

In this section we review other works that relate to different components of our approach, namely: visual localization, scalability, feature learning, and deployment on resource constrained devices.
\PAR{6-DoF visual localization} methods have traditionally been classified as either structure-based or image-based. The former perform direct matching of local descriptors between 2D keypoints of a query image and 3D points in a 3D SfM model~\cite{city-scale-loc, toft2018semantic, active-search, liu2017efficient, inloc}. These methods are able to estimate accurate poses, but often rely on exhaustive matching and are thus compute intensive. As the model grows in size and perceptual aliasing arises, this matching becomes ambiguous, impairing the robustness of the localization, especially under strong appearance changes such as day-night~\cite{loc-benchmark}. Some approaches directly regress the pose from a single image~\cite{brahmbhatt2018geometry, kendall2017geometric}, but are not competitive in term of accuracy~\cite{sattler2019understanding}. Image-based methods are related to image retrieval~\cite{netvlad, torii201524, weyand2016planet} and are only able to provide an approximate pose up to the database discretization, which is not sufficiently precise for many applications~\cite{loc-benchmark, inloc}. They are however significantly more robust than direct local matching as they rely on the global image-wide information. This comes at the cost of increased compute, as state-of-the-art image retrieval is based on large deep learning models.

\PAR{Scalable localization} often deals with the additional compute constrains by using features that are inexpensive to extract, store, and match together~\cite{calonder2010brief, brisk, philbin08lost}. These improve the runtime on mobile devices but further impair the robustness of the localization, limiting their operations to stable conditions~\cite{get-out-of-my-lab}. Hierarchical localization~\cite{irschara, middelberg, corl} takes a different approach by dividing the problem into a global, coarse search followed by a fine pose estimation. Recently,~\cite{corl} proposed to search at the map level using image retrieval and localize by matching hand-crafted local features against retrieved 3D points. As we discuss further in Section~\ref{sec:approach}, its robustness and efficiency are limited by the underlying local descriptors and heterogeneous structure.

\PAR{Learned local features} have recently been developed in attempt to replace hand-crafted descriptors. Dense pixel-wise features naturally emerge from CNNs and provide a powerful representation used for image matching~\cite{ucn, Fathy_2018_ECCV, novotny2018self, rocco2018neighbourhood} and localization~\cite{inloc, loc-benchmark}. Matching dense features is however intractable with limited computing power. 
Sparse learned features, composed of keypoints and descriptors, provide an attractive drop-in replacement to their hand-crafted counterparts and have recently shown outstanding performance~\cite{superpoint, ono2018lf, doap}. They can easily be sampled from dense features, are fast to predict and thus suitable for mobile deployment. CNN keypoint detections have also been shown to outperform classical methods, although they are notably difficult to learn. SuperPoint~\cite{superpoint} learns from self-supervision, while DELF~\cite{delf} employs an attention mechanism to optimize for the landmark recognition task.

\PAR{Deep learning on mobile.} While learning some building blocks of the localization pipeline improves performance and robustness, deploying them on mobile devices is a non-trivial task. Recent advances in multi-task learning allow to efficiently share compute across tasks without manual tuning~\cite{kendallmulti, chen2017gradnorm, sener2018multi}, thus reducing the required network size. Distillation~\cite{distillation} can help to train a smaller network~\cite{mobilenet, zhang2018shufflenet, zoph2018learning} from a larger one that is already trained, but is usually not applied in a multi-task setting.

To the best of our knowledge, our approach is the first of its kind that combines advances in the aforementioned fields to optimize for both efficiency and robustness.
The proposed method seeks to leverage the synergies of these algorithms to deliver a competitive large-scale localization solution and bring this technology closer to real-time, online applications with constrained resources.
\begin{figure}[tb!]
\centering
\includegraphics[width=1.\linewidth]{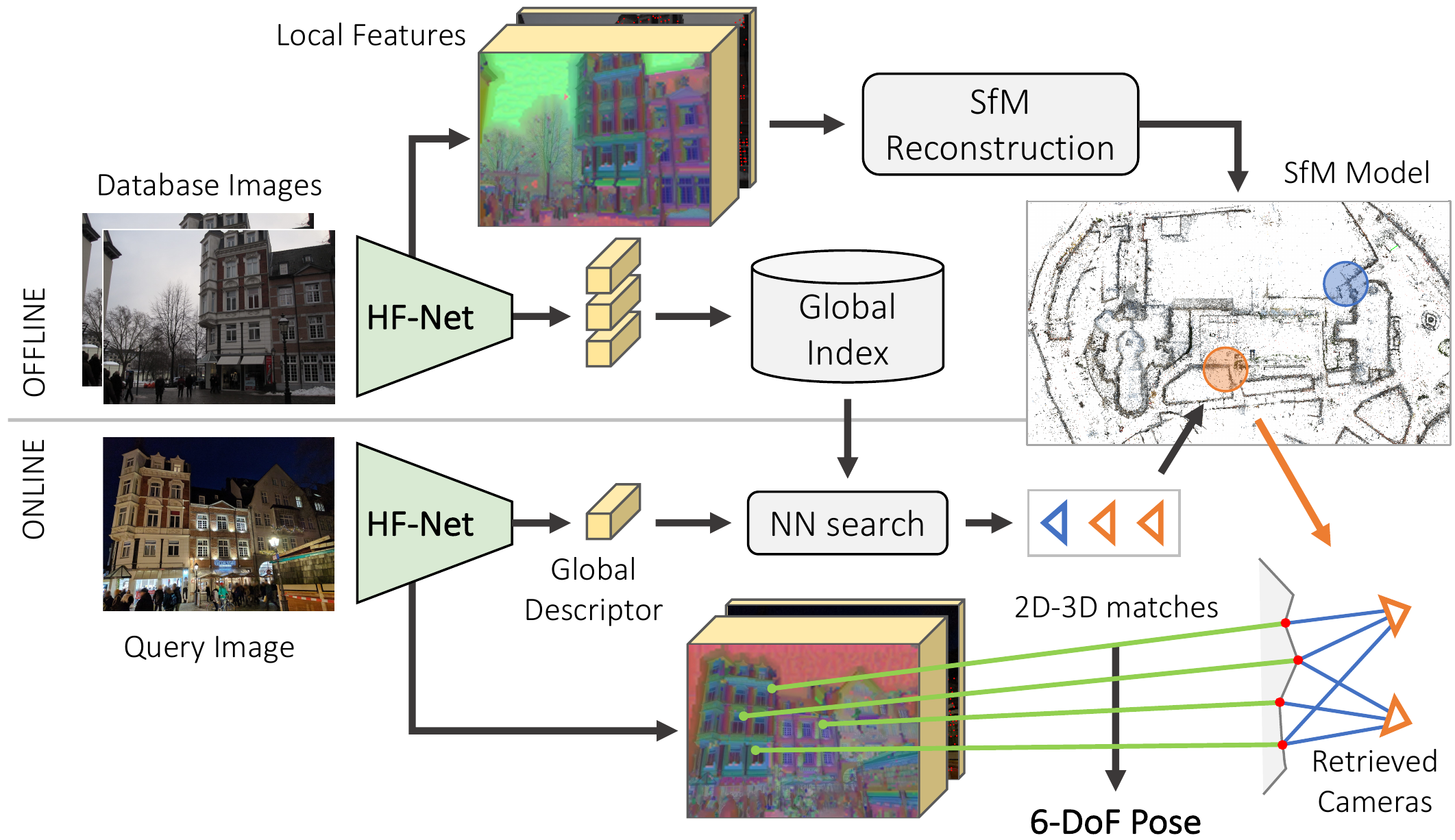}\\
\vspace{1mm}
\caption{\textbf{The hierarchical localization with HF-Net} is significantly simpler than concurrent approaches~\cite{active-search, city-scale-loc}, yet more robust, accurate, and efficient.}
\label{fig:pipeline}
\vspace{-3mm}
\end{figure}

\section{Hierarchical Localization}
\label{sec:approach}
We aim at maximizing the robustness of the localization while retaining tractable computational requirements. Our method is loosely based on the hierarchical localization framework~\cite{corl}, which we summarize here.

\PAR{Prior retrieval.}
A coarse search at the map level is performed by matching the query with the database images using global descriptors. The k-nearest neighbors (NN), called prior frames, represent candidate locations in the map. This search is efficient given that there are far fewer database images than points in the SfM model.

\PAR{Covisibility clustering.}
The prior frames are clustered based on the 3D structure that they co-observe. This amounts to finding connected components, called places, in the covisibility graph that links database images to 3D points in the model.

\PAR{Local feature matching.}
For each place, we successively match the 2D keypoints detected in the query image to the 3D points contained in the place, and attempt to estimate a 6-DoF pose with a PnP~\cite{kneip2011novel} geometric consistency check within a RANSAC scheme~\cite{fischler1981random}. This local search is also efficient as the number of 3D points considered is significantly lower in the place than in the whole model. The algorithm stops as soon as a valid pose is estimated.

\PAR{Discussion.}
In the work of~\cite{corl}, a large state-of-the-art network for image retrieval, NetVLAD~\cite{netvlad}, is distilled into a smaller model, MobileNetVLAD (MNV). This helps to achieve given runtime constraints while partly retaining the accuracy of the original model. The local matching step is however based on SIFT~\cite{lowe2004distinctive}, which is expensive to compute and generates a large number of features, making this step particularly expensive. While this method exhibits good performance in small-scale environments, it does not scale well to larger, denser models. Additionally, SIFT is not competitive with recent learned features, especially under large illumination changes~\cite{doap, ono2018lf, superpoint, mishchuk2017working}. Lastly, a significant part of the computation of local and global descriptors is redundant, as they are both based on the image low-level clues. The heterogeneity of hand-crafted features and CNN image retrieval is thus computationally suboptimal and could be critical on resource-constrained platforms.

\section{Proposed Approach}
We now show how we address these issues and achieve improved robustness, scalability, and efficiency. We first motivate the use of learned features with a homogeneous network structure, and then detail the architecture in Section~\ref{sec:architecture} and our novel training procedure in Section~\ref{sec:training}.

Learned features appear as a natural fit for the hierarchical localization framework. Recent methods like SuperPoint~\cite{superpoint} have shown to outperform popular baseline like SIFT in terms of keypoint repeatability and descriptor matching, which are both critical for localization. Some learned features are additionally significantly sparser than SIFT, thus reducing the number of keypoints to be matched and speeding up the matching step. We show in Section~\ref{exp-feat-eval} that a combination of state-of-the-art networks in image retrieval and local features naturally achieves state-of-the-art localization. This approach particularly excels in extremely challenging conditions, such as night-time queries, outperforming competitive methods by a large margin along with a smaller 3D model size.

While the inference of such networks is significantly faster than computing SIFT on GPU, it still remains a large computational bottleneck for the proposed localization system. With the goal of improving the ability to localize online on mobile devices, we introduce here a novel neural network for hierarchical features, HF-Net, enabling an efficient coarse-to-fine localization. It detects keypoints and computes local and global descriptors in a single shot, thus maximizing sharing of computations, but retaining performance of a larger baseline network. We show in Figure~\ref{fig:pipeline} its application within the hierarchical localization framework.

\begin{figure}[htb]
\begin{center}
\includegraphics[width=1.0\linewidth]{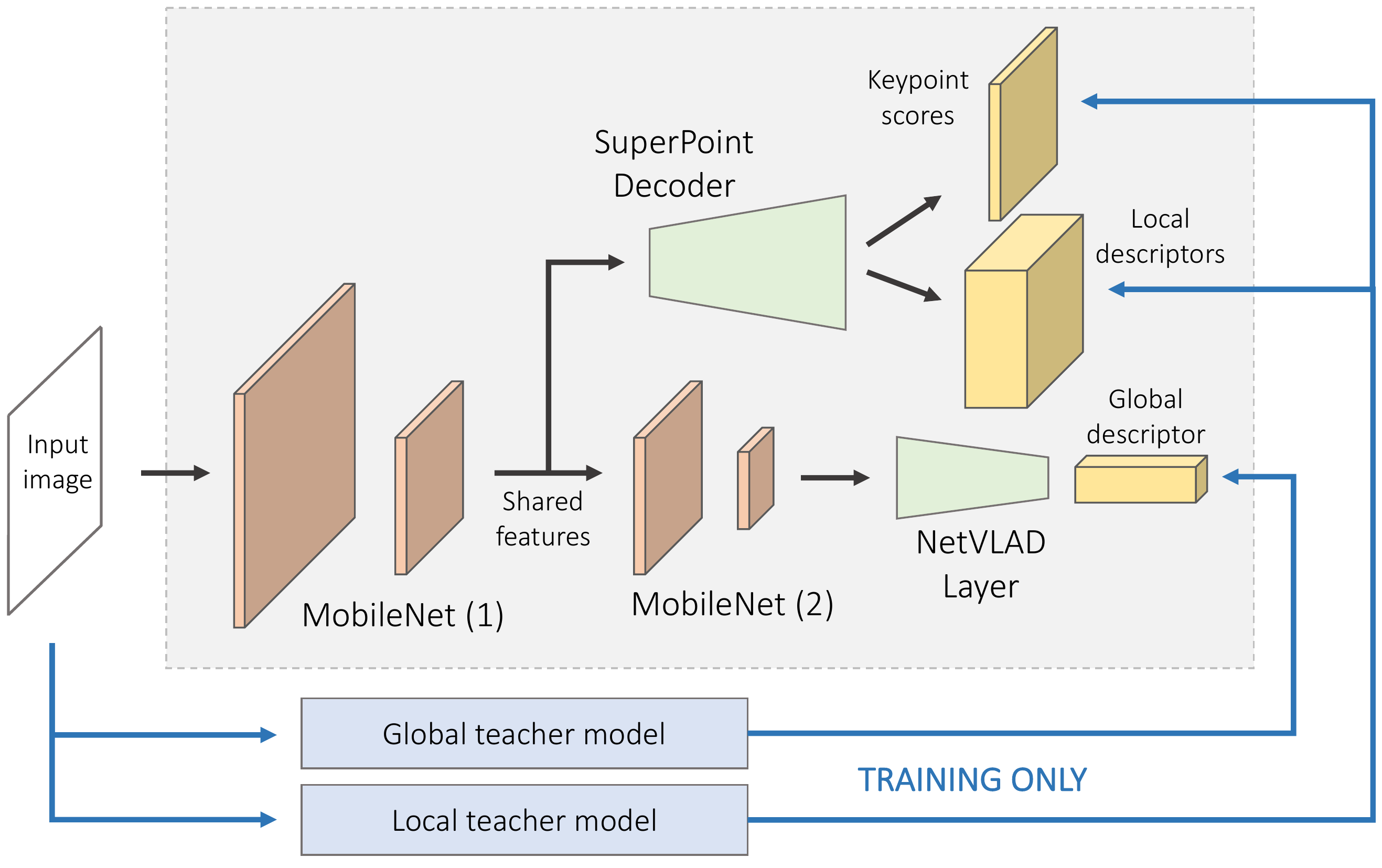}
\end{center}
\vspace{-2mm}
\caption{\textbf{HF-Net} generates three outputs from a single image: a global descriptor, a map of keypoint detection scores, and dense keypoint descriptors. All three heads are trained jointly with multi-task distillation from different teacher networks.}
\label{fig:hfnet}
\vspace{-3mm}
\end{figure}

\subsection{HF-Net Architecture}
\label{sec:architecture}
Convolutional neural networks intrinsically exhibit a hierarchical structure. This paradigm fits well the joint predictions of local and global features and comes at low additional runtime costs. The HF-Net architecture (Figure~\ref{fig:hfnet}) is composed of a single encoder and three heads predicting:
\begin{enumerate*}[label=\roman*\upshape)]
\item keypoint detection scores, \item dense local descriptors and \item a global image-wide descriptor. 
\end{enumerate*}
This sharing of computation is natural: in state-of-the-art image retrieval networks, the global descriptors are usually computed from the aggregation of local feature maps, which might be useful to predict local features.

The encoder of HF-Net is a MobileNet~\cite{mobilenet} backbone, a popular architecture optimized for mobile inference. Similarly to MNV~\cite{corl}, the global descriptor is computed by a NetVLAD layer~\cite{netvlad} on top of the last feature map of MobileNet. For the local features, the SuperPoint~\cite{superpoint} architecture is appealing for its efficiency, as it decodes the keypoints and local descriptors in a fixed non-learned manner. This is much faster than applying transposed convolutions to upsample the features.
It predicts dense descriptors which are fast to sample bilinearly, resulting in a runtime independent from the number of detected keypoints. 
On the other hand, patch-based architectures like LF-Net~\cite{ono2018lf} apply a Siamese network to image patches centered at all keypoint locations, resulting in a computational cost proportional to the number of detections.

For its efficiency and flexibility, we thus adopt the SuperPoint decoding scheme for keypoints and local descriptors. The local feature heads branch out from the MobileNet encoder at an earlier stage than the global head, as a higher spatial resolution is required to retain spatially discriminative features, local features are on a lower semantic level than image-wide descriptors~\cite{Fathy_2018_ECCV}.

\subsection{Training Process}
\label{sec:training}

\PAR{Data scarcity.} Local and global descriptors are often trained with metric learning using ground truth positive and negative pairs of local patches and full images. These ground truth correspondences are particularly difficult to obtain at the scale required to train large CNNs. While global supervision naturally emerges from local correspondences, there is currently no such dataset that simultaneously 
\begin{enumerate*}[label=\roman*\upshape)]
\item exhibits a sufficient perceptual diversity at the global image level, \eg with various conditions such as day, night, seasons, and 
\item contains ground truth local correspondences between matching images. 
\end{enumerate*}
These correspondences are often recovered from the dense depth~\cite{ono2018lf} computed from an SfM model~\cite{schoenberger2016sfm, schoenberger2016mvs}, which is intractable to build at the scale required by image retrieval.

\PAR{Data augmentation.} Self-supervised methods that do not rely on correspondences, such as SuperPoint, require heavy data augmentation, which is key to the invariance of the local descriptor. While data augmentation often captures well the variations in the real world at the local level, it can break the global consistency of the image and make the learning of the global descriptor very challenging.

\PAR{Multi-task distillation} is our solution to this data problem. We employ distillation to learn the representation directly from an off-the-shelf trained teacher model. This alleviates the above issues, with a simpler and more flexible training setup that allows the use of arbitrary datasets, as infinite amount of labeled data can be obtained from the inference of the teacher network. Directly learning to predict the output of the teacher network additionally eases the learning task, allowing to directly train a smaller student network. We note an interesting similarity with SuperPoint, whose detector is training by bootstrapping, supervised by itself through the different training runs. This process could also be referred as self-distillation, and shows the effectiveness of distillation as a practical training scheme.

The supervision of local and global features can originate from different teacher networks, resulting in a multi-task distillation training that allows to leverage state-of-the-art teachers. Recent advances~\cite{kendallmulti} in multi-task learning enable a student $s$ to optimally copy all teachers $t_{1,2,3}$ without any manual tuning of the weights that balance the loss:
\begin{equation}
\begin{split}
    L &= e^{-w_1}||\*d^g_{s} - \*d^g_{t_1}||_2^2 
    + e^{-w_2}||\*d^l_{s} - \*d^l_{t_2}||_2^2\\
    &+ 2e^{-w_3}\mathrm{CrossEntropy}(\*p_{s}, \*p_{t_3})
    + \sum_{i} w_i,
\end{split}
\end{equation}%
where $\*d^g$ and $\*d^l$ are global and local descriptors, $\*p$ are keypoint scores, and $w_{1,2,3}$ are optimized variables.

More generally, our formulation of the multi-task distillation can be applied to any application that requires multiple predictions while remaining computationally efficient, particularly in settings where ground truth data for all tasks is expensive to collect. It could also be applied to some hand-crafted descriptors deemed too compute-intensive.

\section{Experiments}

In this section, we present experimental evaluations of the building blocks of HF-Net and of the network as a whole.
We want to prove its applicability to large-scale localization problems in challenging conditions while remaining computationally tractable.
We first perform in Section~\ref{exp-feat-eval} a thorough evaluation of current top-performing classical and learning-based methods for local feature detection and description.
Our goal is to explain how these insights influenced the design choices of HF-Net presented in Section~\ref{exp-impl}.
We then evaluate in Section~\ref{exp-loc} our method on challenging large-scale localization benchmarks~\cite{loc-benchmark} and demonstrate the advantages of the coarse-to-fine localization paradigm.
To address our real-time localization focus, we conclude with runtime considerations in Section~\ref{exp-timing}. 

\subsection{Local Features Evaluation}
\label{exp-feat-eval}
We start our evaluation by investigating the performance of local matching methods under different settings on two datasets, HPatches~\cite{balntas20017hpatches} and SfM~\cite{ono2018lf}, that provide dense ground truth correspondences between image pairs for both 2D and 3D scenes.

\PAR{Datasets.}
HPatches~\cite{balntas20017hpatches} contains 116 planar scenes containing illumination and viewpoint changes with 5 image pairs per scene and ground truth homographies. 
SfM is a dataset built by~\cite{ono2018lf} composed of photo-tourism collections collected by~\cite{heinly2015reconstructing, thomee2015yfcc100m}. Ground truth correspondences are obtained from dense per-image depth maps and relative 6-DoF poses, computed using COLMAP~\cite{schoenberger2016sfm}. We select 10 sequences for our evaluation and for each randomly sample 50 image pairs with a given minimum overlap. A metric scale cannot be recovered with SfM reconstruction but is important to compute localization metrics. We therefore manually label each SfM model using metric distances measured in Google Maps.

\PAR{Metrics.} We compute and aggregate pairwise metrics defined by~\cite{superpoint} over all pairs for each dataset. For the detectors, we report the repeatability and localization error of the keypoint locations. Both are important for visual localization as they can impact the number of inlier matches, the reliability of the matches, but also the quality of the 3D model. We compute nearest neighbor matches between descriptors and report the mean average precision and the matching score. The former reflects the ability of the method to reject spurious matches. The latter assesses the quality of the detector and the descriptor together. We also compute the recall of pose estimation, either a homography for HPatches or a 6-DoF pose for the SfM dataset, with thresholds of 3~pixels and 3 meters, respectively.

\PAR{Methods.} We evaluate the classical detectors Difference of Gaussian (DoG) and Harris~\cite{harris1988combined} and the descriptor RootSIFT~\cite{arandjelovic2012three}. For the learning-based methods, we evaluate the detections and descriptors of SuperPoint~\cite{superpoint} and LF-Net~\cite{netvlad-tf}. We additionally evaluate a dense version of DOAP~\cite{doap} and the feature map \verb|conv3_3| of NetVLAD~\cite{netvlad} and use SuperPoint detections for both. More details are provided in the appendix.

\PAR{Detectors.} We report the results in Table~\ref{tab:kpts_results:comparison}. Harris exhibits the highest repeatability but also the highest localization error. Conversely, DoG is less repeatable but has the lowest error, likely due to the multi-scale detection and pixel refinement. SuperPoint seems to show the best trade-off between repeatability and error.

\begin{table}[htb]
\begin{center}
\scriptsize{
\setlength\tabcolsep{3.0pt}
\begin{tabular}{l|cc|cc}
\toprule
& \multicolumn{2}{c|}{\textbf{HPatches}}& \multicolumn{2}{c}{\textbf{SfM}}\\
& {Rep.}& {MLE}& {Rep.}& {MLE}\\
\midrule
{DoG}        & 0.307 & 0.94 & 0.284 & 1.20 \\
{Harris}     & 0.535 & 1.14 & 0.510 & 1.46 \\
{SuperPoint} & 0.495 & 1.04 & 0.509 & 1.45 \\
{LF-Net}     & 0.460 & 1.13 & 0.454 & 1.44 \\
\bottomrule
\end{tabular}

}
\end{center}
\vspace{-6pt}
\caption{\textbf{Evaluation of the keypoint detectors.} We report the repeatability (rep.) and mean localization error (MLE).}
\label{tab:kpts_results:comparison}
\end{table}

\begin{table}[htb!]
\begin{center}
\scriptsize{
\setlength\tabcolsep{3.0pt}
\begin{tabular}{l|ccc|ccc}
\toprule
& \multicolumn{3}{c|}{\textbf{HPatches}}& \multicolumn{3}{c}{\textbf{SfM}}\\
(detector / descriptors) & {Homography}& {MS} & {mAP} & {Pose} & {MS} & {mAP}\\
\midrule
{Root-SIFT} & 0.681 & 0.307 & 0.651 & 0.700 & 0.199 & 0.236 \\
{LF-Net} & 0.629 & 0.305 & 0.572 & 0.676 & 0.221 & 0.207 \\
{SuperPoint}       & 0.810 & 0.441 & 0.846 & 0.794 & 0.418 & 0.488 \\
{Harris / SuperPoint}       & 0.669 & 0.448 & 0.737 & 0.684 & 0.404 & 0.397 \\
{SuperPoint / DOAP}     & - & - & - & 0.838  & 0.448 & 0.554\\
{SuperPoint / NetVLAD}     & 0.788 & 0.419 & 0.798 & 0.800 & 0.374 & 0.423 \\
\bottomrule
\end{tabular}

}
\end{center}
\vspace{-6pt}
\caption{\textbf{Evaluation of the local descriptors.} The matching score (MS) and mean Average Precision (mAP) are reported, in addition to the homography correctness for HPatches and the pose accuracy for the SfM dataset.}
\label{tab:desc_results:comparison}
\vspace{-4mm}
\end{table}

\PAR{Descriptors.} DOAP outperforms SuperPoint on all metrics on the SfM dataset, but cannot be evaluated on HPatches as it was trained on this dataset. NetVLAD shows good pose estimation but poor matching precision on SfM, which is disadvantageous when the number of keypoints is limited or the inlier ratio important, \eg for localization. Overall, it stands that learned features outperform hand-crafted ones. 

Interestingly, SuperPoint descriptors perform poorly when extracted from Harris detections, although the latter is also a corner detector with high repeatability. This hints that learned descriptors can be highly coupled with the corresponding detections.

LF-Net and SIFT, both multi-scale approaches with subpixel detection and patch-based description, are outperformed by dense descriptors like DOAP and SuperPoint. A simple representation trained with the right supervision can thus be more effective than a complex and computational-heavy architecture. We note that SuperPoint requires significantly fewer keypoints to estimate a decent pose, which is highly beneficial for runtime-sensitive applications.
\subsection{Implementation Details}
\label{exp-impl}

Motivated by the results presented in Section~\ref{exp-feat-eval}, this section briefly introduces the design and implementation of HF-Net. Below, we explain our choices of the distillation teacher models, training datasets and improvements to the baseline 2D-3D local matching.

\PAR{Teacher models.}
We evaluate the impact of the two best descriptors, DOAP and SuperPoint, on the localization in Section~\ref{exp-loc}.
Results show that the latter is more robust to day-night appearance variations, as its training set included low-light data. We eventually chose it as the supervisor teacher network for the descriptor head of HF-Net. The global head is supervised by NetVLAD.

\PAR{Training data.} In this work, we target urban environments in both day and night conditions. To maximize the performance of the student model on this data, we select training data that fits this distribution. We thus train on 185k images from the Google Landmarks dataset~\cite{delf}, containing a wide variety of day-time urban scenes, and 37k images from the night and dawn sequences of the Berkeley Deep Drive dataset~\cite{bdd}, composed of road scenes with motion blur. We found the inclusion of night images in the training dataset to be critical for the generalization of the global retrieval head to night queries. For example, a network trained on day-time images only would easily confuse a night-time dark sky with a day-time dark tree. We also train with photometric data augmentation but use the targets predicted on the clean images.

\PAR{Efficient hierarchical localization.} Sarlin \etal~\cite{corl} identified the local 2D-3D matching as the bottleneck of the pipeline. Our system significantly improves on the efficiency of their approach:
\begin{enumerate*}[label=\roman*)]
\item Spurious local matches are filtered out using a modified ratio test that only applies if the first and second nearest neighbor descriptors correspond to observations of different 3D points, similarly to~\cite{mishkin2015mods}, thus retaining more matches in highly covisible areas.
\item Learned global and local descriptors are normalized and matched with a single matrix multiplication on GPU.
\end{enumerate*}
Additional implementation details and hyperparameters are provided in the appendix.
\subsection{Large-scale Localization}
\label{exp-loc}

\begin{table*}[htb!]
\begin{center}
\scriptsize{
\setlength\tabcolsep{1.1pt}

\begin{tabular}{l|c|c||c|c|c|c||c|c}& \multicolumn{2}{c||}{\textbf{Aachen}}& \multicolumn{4}{c||}{\textbf{RobotCar}}& \multicolumn{2}{c}{\textbf{CMU}}\\

& {day}& {night}& {dusk}& {sun}& {night}& {night-rain}& {urban}& {suburban}\\ \cline{2-9}

\multicolumn{1}{r|}{\begin{tabular}[c]{@{}r@{}}distance [m]\\ orient. [deg]\end{tabular}} & {\begin{tabular}[c]{@{}c@{}}.25/.50/5.0\\ 2/5/10\end{tabular}} & {\begin{tabular}[c]{@{}c@{}}0.5/1.0/5.0\\ 2/5/10\end{tabular}} & {\begin{tabular}[c]{@{}c@{}}.25/.50/5.0\\ 2/5/10\end{tabular}} & {\begin{tabular}[c]{@{}c@{}}.25/.50/5.0\\ 2/5/10\end{tabular}} & {\begin{tabular}[c]{@{}c@{}}.25/.50/5.0\\ 2/5/10\end{tabular}} &
{\begin{tabular}[c]{@{}c@{}}.25/.50/5.0\\ 2/5/10\end{tabular}} & {\begin{tabular}[c]{@{}c@{}}.25/.50/5.0\\ 2/5/10\end{tabular}} &{\begin{tabular}[c]{@{}c@{}}.25/.50/5.0\\ 2/5/10\end{tabular}}\\ \hline

{AS} & 57.3 / 83.7 / \hlf{96.6} & 19.4 / 30.6 / 43.9 & 44.7 / 74.6 / \hls{95.9} & 25.0 / 46.5 / 69.1 & \00.5 / \01.1 / \03.4\0 & \01.4 / \03.0 / \05.2\0 &  55.2 / 60.3 / 65.1 & 20.7 / 25.9 / 29.9 \\ \hline

{CSL} & 52.3 / 80.0 / \hls{94.3} & 24.5 / 33.7 / 49.0 & \hlf{56.6} / 82.7 / \hls{95.9} & 28.0 / 47.0 / 70.4 & \00.2 / \00.9 / \05.3\0 & \00.9 / \04.3 / \09.1\0 & 36.7 / 42.0 / 53.1 & \08.6 / 11.7 / 21.1\\ \hline

{DenseVLAD} & \00.0 / \00.1 / 22.8 & \00.0 / \02.0 / 14.3 & 10.2 / 38.8 / 94.2 & \05.7 / 16.3 / 80.2 & \00.9 / \03.4 / 19.9\0  & \01.1 / \05.5 / \hls{25.5}\0 & 22.2 / 48.7 / 92.8 & \09.9 / 26.6 / 85.2\\ \hline

{NetVLAD} & \00.0 / \00.2 / 18.9 & \00.0 / \02.0 / 12.2 & \07.4 / 29.7 / 92.9 & \05.7 / 16.5 / 86.7 & \00.2 / \01.8 / 15.5\0 & \00.5 / \02.7 / 16.4\0 & 17.4 / 40.3 / 93.2 & \07.7 / 21.0 / 80.5\\ \hline

{SMC} & - & - & (53.8 / 83.0 / 97.7) & (46.7 / 74.6 / 95.9) & (6.2 / 18.5 / 44.3) & (8.0 / 26.4 / 46.4) & 75.0 / 82.1 / 87.8 & 44.0 / 53.6 / 63.7 \\ \hline

{NV+SIFT} & \hlf{82.8} / \hlf{88.1} / 93.1 & 30.6 / 43.9 / 58.2 & \hls{55.6} / \hlf{83.5} / 95.3 & 46.3 / 67.4 / \hls{90.9} & \0\hls{4.1} / \0\hls{9.1} / \hls{24.4}\0 & \02.3 / 10.2 / 20.5\0 & 63.9 / 71.9 / 92.8 & 28.7 / 39.0 / 82.1\\ \hline

{\textbf{NV+SP (ours)}\0} & \hls{79.7} / \hls{88.0} / 93.7 & \hlf{40.8} / \hlf{56.1} / \hlf{74.5} & 54.8 / \hls{83.0} / \hlf{96.2} & \hlf{51.7} / \hlf{73.9} / \hlf{92.4} & \0\hlf{6.6} / \hlf{17.1} / \hlf{32.2}\0 & \0\hlf{5.2} / \hlf{17.0} / \hlf{26.6}\0 & \hlf{91.7} / \hlf{94.6} / \hlf{97.7} & \hlf{74.6} / \hlf{81.6} / \hlf{91.4}\\ \hline

{\textbf{HF-Net (ours)}\0} & 75.7 / 84.3 / 90.9 & \hlf{40.8} / \hls{55.1} / \hls{72.4} & 53.9 / 81.5 / 94.2 & \hls{48.5} / \hls{69.1} / 85.7 & \02.7 / \06.6 / 15.8\0 & \0\hls{4.7} / \hls{16.8} / 21.8\0 & \hls{90.4} / \hls{93.1} / \hls{96.1} & \hls{71.8} / \hls{78.2} / \hls{87.1}\\ \hline

\end{tabular}

}
\end{center}
\vspace{-6pt}
\caption{\textbf{Evaluation of the localization} on the Aachen Day-Night, RobotCar Seasons, and CMU Seasons datasets. We report the recall~[\%] at different distance and orientation thresholds and highlight for each of them the \hlf{best} and \hls{second-best} methods. X+Y denotes hierarchical localization with X (Y) as global (local) descriptors. SMC is excluded from the comparison for RobotCar as it uses extra semantic data.
}
\vspace{-1mm}
\label{tab:loc_results:comparison}
\end{table*}

Under the light of the local evaluation, we now evaluate our hierarchical localization on three challenging large-scale benchmarks introduced by~\cite{loc-benchmark}.

\PAR{Datasets.} Each dataset is composed of a sparse SfM model built with a set of reference images. The Aachen Day-Night dataset~\cite{sattler2012image} contains 4,328 day-time database images from a European old town, and 824 and 98 queries taken in day and night conditions respectively. The RobotCar Seasons dataset~\cite{maddern2017robotcar} is a long-term urban road dataset that spans multiple city blocks. It is composed of 20,862 overcast reference images and a total of 11,934 query images taken in multiple conditions, such as sun, dusk, and night. Lastly, the CMU Seasons dataset~\cite{bansal2014localization} was recorded in urban and suburban environments over a course of 8.5~km. It contains 7,159 reference images and 75,335 query images recorded in different seasons. This dataset is of significantly lower scale as the queries are localized against isolated submodels containing around 400 images each.

\PAR{Large scale model construction.} SfM models built with COLMAP~\cite{schoenberger2016sfm, schoenberger2016mvs} using RootSIFT are provided by the dataset authors. These are however not suitable when localizing with methods based on different feature detectors. We thus build new 3D models with keypoints detected by SuperPoint and HF-Net. The process is as follows: 
\begin{enumerate*}[label=\roman*\upshape)]
\item we perform 2D-2D matching between reference frames using our features and an initial filtering ratio test;
\item the matches are further filtered within COLMAP using two-view geometry;
\item 3D points are triangulated using the provided ground truth reference poses. 
\end{enumerate*}
Those steps result in a 3D model with the same scale and reference frame as the original one.

\PAR{Comparison of model quality.} The HF-Net Aachen model contains fewer 3D points (685k vs 1,899k for SIFT) and fewer 2D keypoints per image (2,576 vs 10,230 for SIFT). However, a larger ratio of the original 2D keypoints is matched (33.8\% vs 18.8\% for SIFT), and each 3D point is on average observed from more reference images. Matching a query keypoint against this model is thus more likely to succeed, showing that our feature network produces 3D models more suitable for localization.

\begin{figure*}[htb]
\centering
\begin{minipage}{.3\linewidth}
    \centering
    \includegraphics[width=0.9\linewidth]{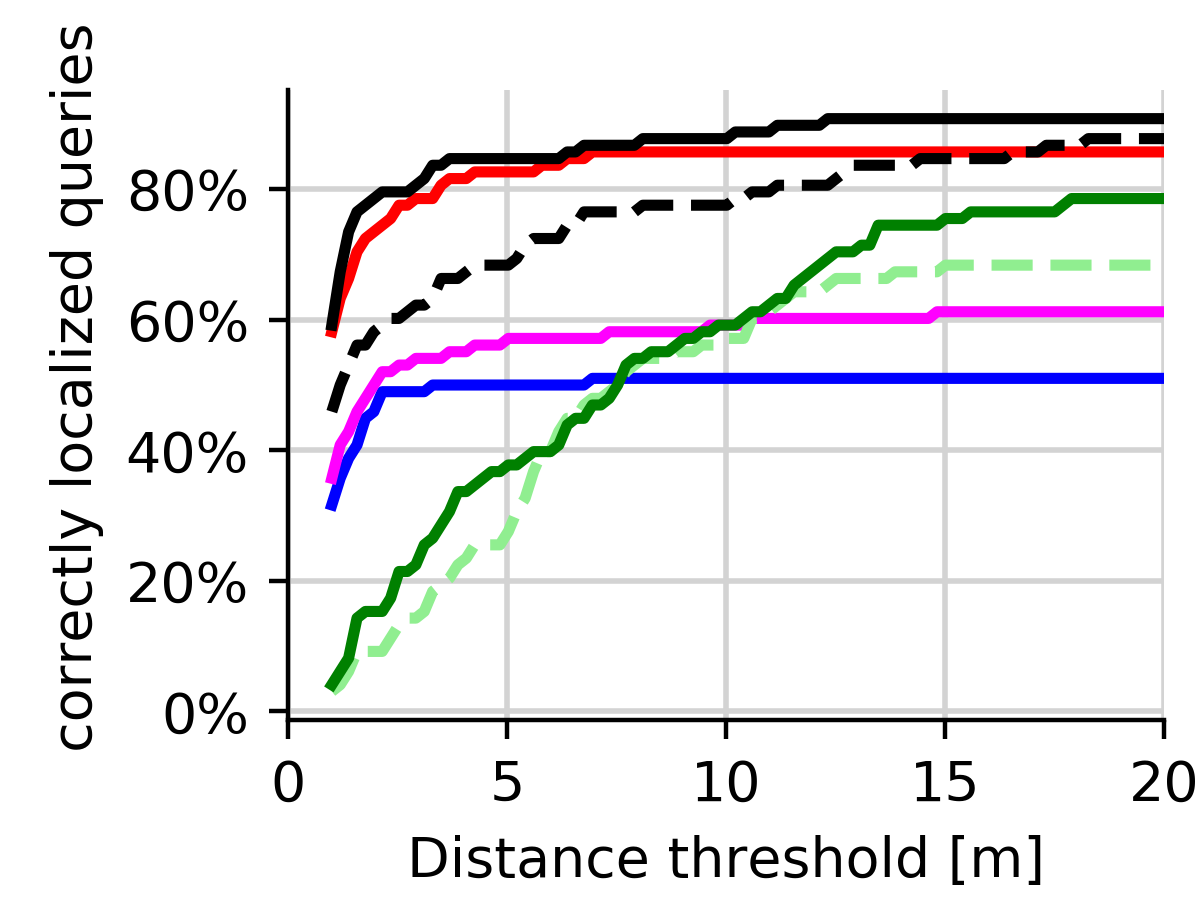}
\end{minipage}%
\begin{minipage}{0.3\linewidth}
    \centering
    \includegraphics[width=0.9\linewidth]{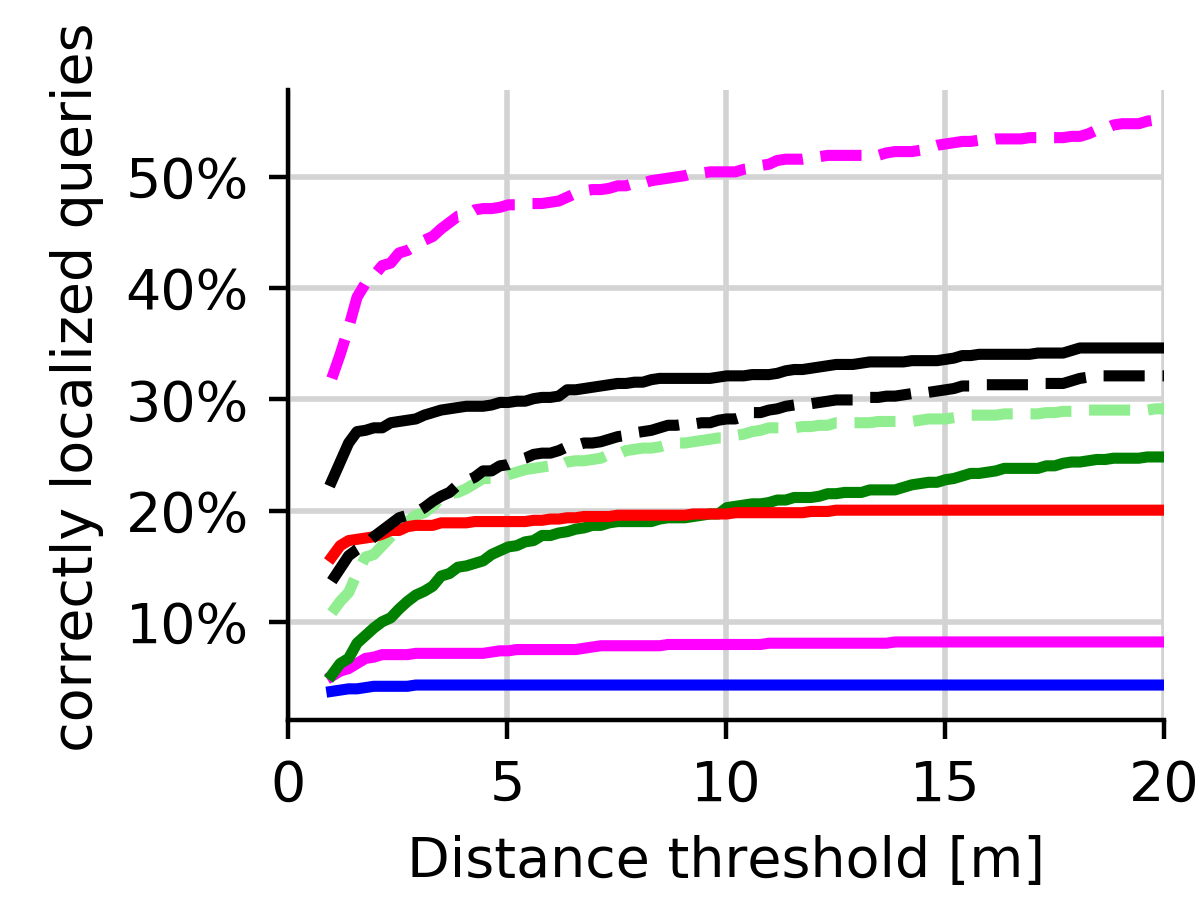}
\end{minipage}%
\begin{minipage}{0.3\linewidth}
    \centering
    \includegraphics[width=0.9\linewidth]{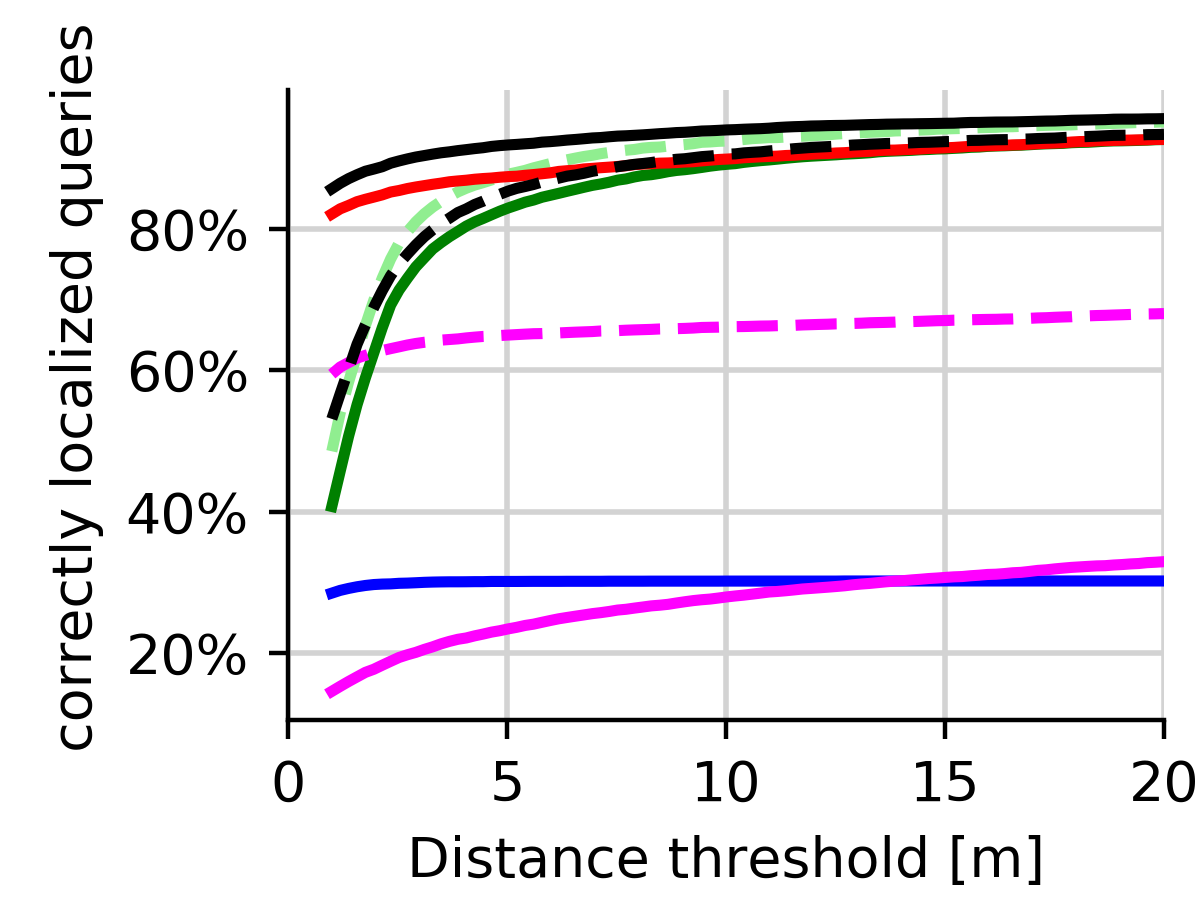}
\end{minipage}\\
\vspace{1mm}
\includegraphics[height=3mm]{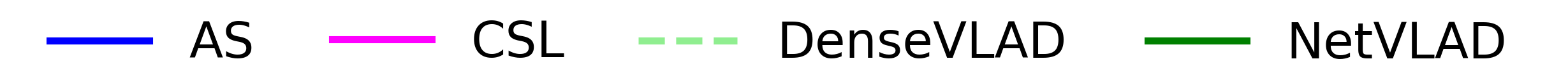}
\includegraphics[height=3mm]{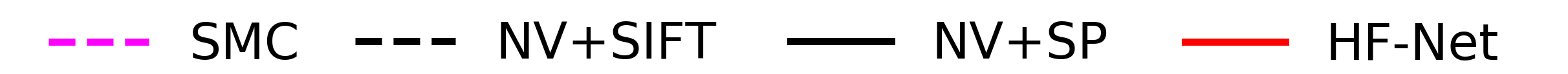}\\
\vspace{1mm}
\caption{\textbf{Cumulative distribution of position errors} for the Aachen night (left), RobotCar night-all (center) and CMU suburban (right) datasets. 
On Aachen, HF-Net and NV+SP have similar performance and outperform approaches based on global retrieval and on feature matching.
On RobotCar, HF+Net performs worse than NV+SP, which suggests a limitation of the distillation process.
On CMU, the hierarchical localization shows a significant boost over other methods, particularly for small distance thresholds.}
\label{fig:cumulative}
\vspace{-3mm}
\end{figure*}

\PAR{Methods.} We first evaluate our hierarchical localization based on learned features extracted by NetVLAD~\cite{netvlad} and SuperPoint~\cite{superpoint}. Named NV+SP, it uses the most powerful predictors available. We then evaluate a more efficient localization with global descriptors and local features computed by HF-Net. We also consider several localization baselines evaluated by the benchmark authors. Active Search~(AS)~\cite{active-search} and City Scale Localization~(CSL)~\cite{city-scale-loc} are both 2D-3D direct matching methods representing the current state-of-the-art in terms of accuracy. DenseVLAD~\cite{torii201524} and NetVLAD~\cite{netvlad} are image retrieval approaches that approximate the pose of the query by the pose of the top retrieved database image. The recently-introduced Semantic Match Consistency (SMC)~\cite{toft2018semantic} relies on semantic segmentation for outlier rejection. It assumes known gravity direction and camera height and, for the RobotCar dataset, was trained on the evaluation data using ground truth semantic labels. We introduce an additional baseline, NV+SIFT, that performs hierarchical localization with RootSIFT as local features, and is an upper bound to the MNV+SIFT method of~\cite{corl}.

\PAR{Results.} We report the pose recall at position and orientation thresholds different for each sequence, as defined by the benchmark~\cite{loc-benchmark}. Table~\ref{tab:loc_results:comparison} shows the localization results for the different methods. Cumulative plots for the three most challenging sequences are presented in Figure~\ref{fig:cumulative}.

\PAR{Localization with NV+SP.} On the Aachen dataset, NV+SP is competitive on day-time queries and outperforms all methods for night-time queries, where the performance drop w.r.t.\ the day is significantly smaller than for direct matching methods, which suffer from the increased ambiguity of the matches.
On the RobotCar dataset, it performs similarly to other methods on the dusk sequence, where the accuracy tends to saturate. In the more challenging sequences, image retrieval methods tend to work better than direct matching approaches, but are far outperformed by NV+SP in both fine-\ and coarse-precision regimes.
On the difficult CMU dataset, NV+SP achieves an outstanding robustness compared to all baselines, including the most recent SMC.
Overall, NV+SP sets a new state-of-the-art on the CMU dataset and on the challenging sequences of the Aachen and RobotCar datasets. The superior performance in both fine-\ and coarse-precision regimes shows that our approach is both more accurate and more robust.

\PAR{Comparison with NV+SIFT.} We observe that NV+SIFT consistently outperforms AS and CSL, although all methods are based on the same RootSIFT features. This shows that our hierarchical approach with a coarse initial prior brings significant benefits, especially in challenging conditions where image-wide information helps disambiguate matches. It thus provides a better outlier rejection than complex domain-specific heuristics used in AS and CSL. The superiority of NV+SP highlights the simple gain of learned features like SuperPoint. On the Aachen night and RobotCar dusk sequences, which are the easiest ones, NV+SIFT performs marginally better than NV+SP for the fine threshold. This is likely due to the lower localization accuracy of the SuperPoint keypoints, as highlighted in Section~\ref{exp-feat-eval}, since DoG performs a subpixel refinement.

\PAR{Localization with HF-Net.} On most sequences, HF-Net performs similarly to its upper bound NV+SP, with a recall drop of 2.6\% on average. We show qualitative results in Figure~\ref{fig:qual} and in the appendix. In the RobotCar night sequences, HF-Net is significantly worse than NV+SP. We attribute this to the poor performance of the distilled global descriptors on blurry low-quality images. This highlights a clear limitation of our approach: on large, self-similar environment, the model capacity of HF-Net becomes the limiting factor. A complete failure of the global retrieval directly translates into a failure of the hierarchical localization.

\begin{table}[htb!]
\begin{center}
\scriptsize{
\setlength\tabcolsep{3.0pt}
\begin{tabular}{l|c|c|c|c|c}
                       & Distance thresh. & NV+SP & NV+HF-Net & NV+DOAP & HF-Net   \\ \hline
\multirow{3}{*}{Day}   & 0.25m  & 79.7  & 81.2  & 80.0    & 75.7 \\
                       & 0.5m   & 88.0  & 88.2  & 88.5    & 84.3 \\
                       & 5m     & 93.7  & 94.2  & 93.3    & 90.9 \\ \hline
\multirow{3}{*}{Night} & 0.5m   & 40.8  & 40.8  & 34.7    & 40.8 \\
                       & 1m     & 56.1  & 56.1  & 52.0    & 55.1 \\
                       & 5m     & 74.5  & 76.5  & 72.4    & 72.4
\end{tabular}
}
\end{center}
\vspace{-6pt}
\caption{\textbf{Ablation study} on the Aachen Day-Night dataset. We report the recall~[\%] of the hierarchical localization with diffrent gloabal descriptors (NetVLAD and HF-Net) and local features (SuperPoint, DOAP, and HF-Net).}
\label{tab:ablation}
\vspace{-3mm}
\end{table}

\PAR{Ablation study.} In Table~\ref{tab:ablation}, we evaluate the influence of different predictors within the hierarchical localization framework. Comparing NV+SP with NV+HF, we note that local HF-Net features perform better than the SuperPoint model that was used to train them. This demonstrates the benefits of multi-task distillation, where the supervision signal from the global teacher can improve intermediate features and help local descriptors. We also observe that the localization with DOAP is significantly worse at night, which might be due to the complex augmentation schemes SuperPoint is based on. Finally, the comparison of HF-Net with NV+HF-Net reveals that HF-Net global descriptors have a somewhat limited capacity compared to the original NetVLAD and are limiting the performance.

\begin{figure}[tb!]
\centering
\includegraphics[trim=0 140 0 0,clip,width=\linewidth]{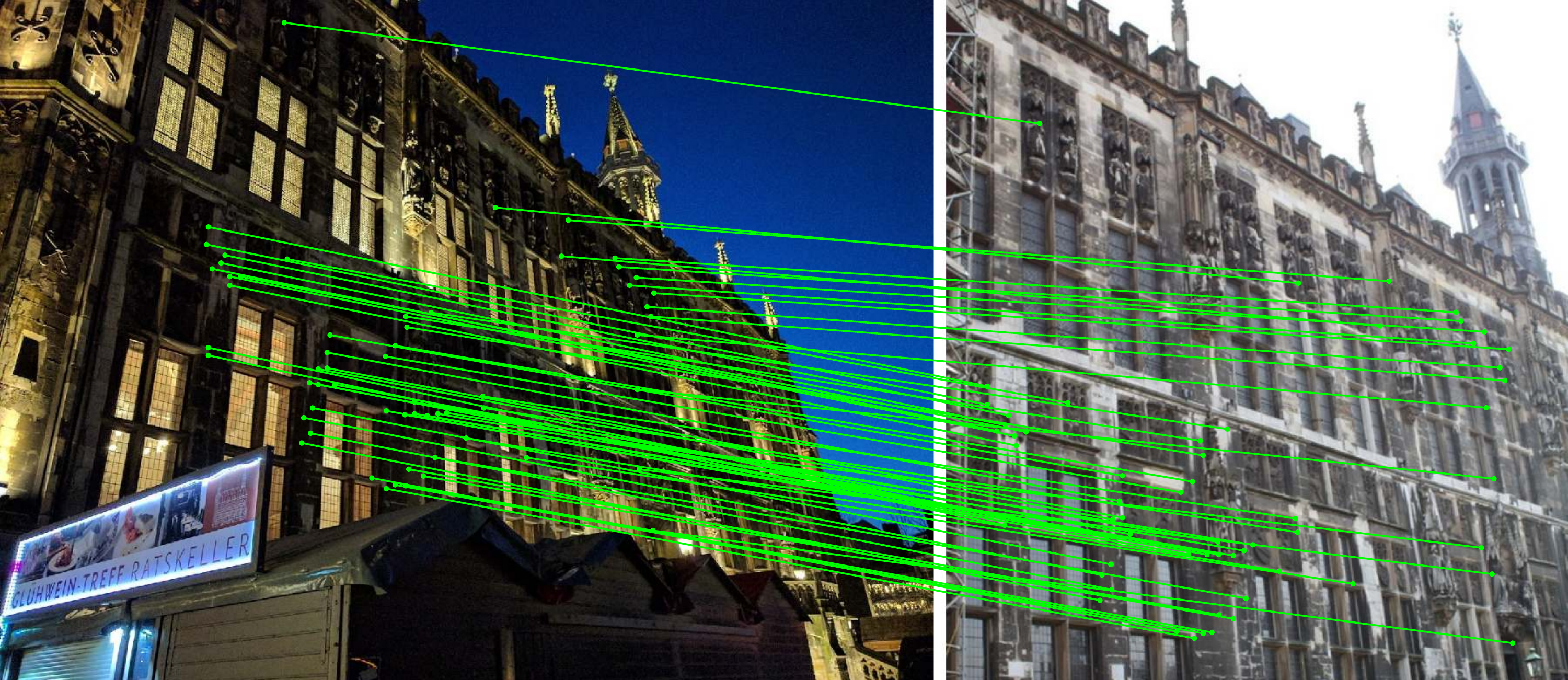}\\[0.5mm]
\includegraphics[trim=0 35 0 0,clip,width=\linewidth]{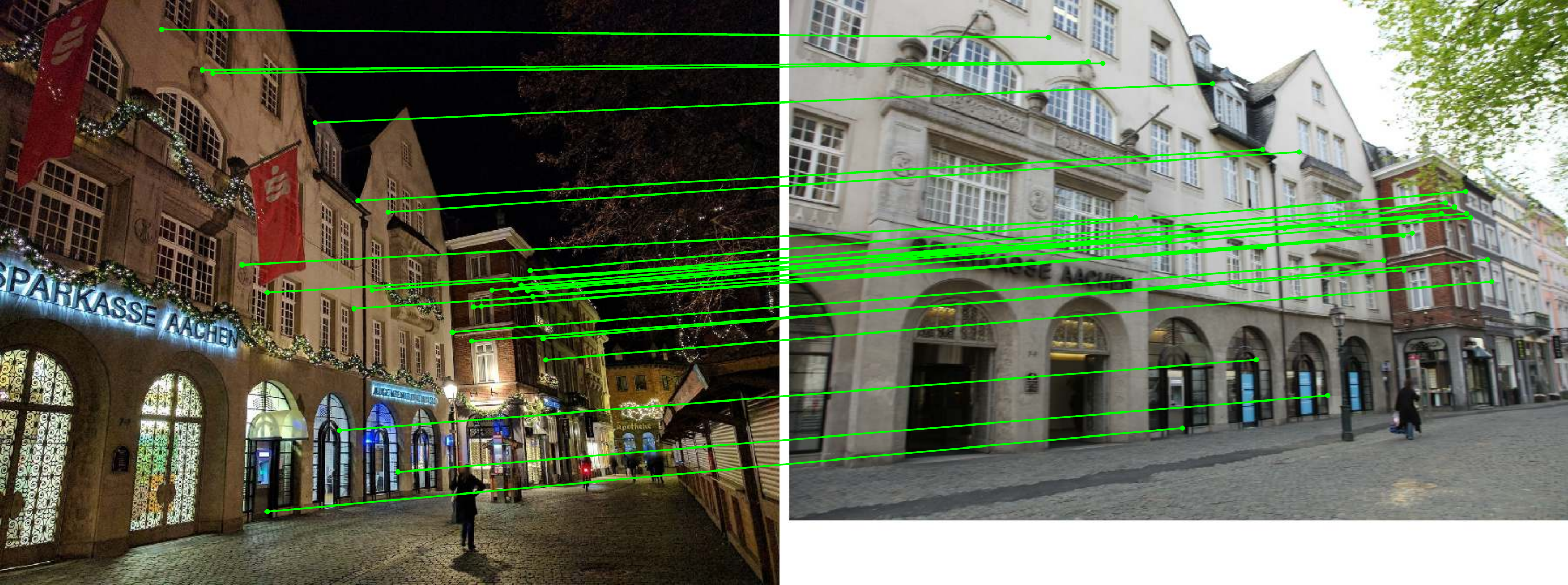}
\caption{\textbf{Successful localization with HF-Net} on the Aachen Day-Night dataset. We show two queries (left) and the retrieved database images with the most inlier matches (right).}
\vspace{-4mm}
\label{fig:qual}
\end{figure}
\subsection{Runtime Evaluation}
\label{exp-timing}

As our propose localization solution was developed keeping the computational constraints in mind, we analyze its runtime and compare it with baselines presented in Section~\ref{exp-loc}. These were measured on a PC equipped with an Intel Core i7-7820X CPU (3.60GHz) CPU, 32GB of RAM and an NVIDIA GeForce GTX 1080 GPU. Table~\ref{tab:timings} presents the detailed timings.

\begin{table}[hb!]
\begin{center}
\scriptsize{
\resizebox{\columnwidth}{!}{%
\setlength\tabcolsep{3.0pt}
\newcommand\nicehrule{\leaders\hbox{\rule[0.3em]{.1pt}{0.4pt}}\hfill\mbox{}}
\newcommand\hcellrule[2]{\multicolumn{#1}{c|}{\nicehrule\ #2\ \nicehrule}}

\begin{tabular}{c|c|c|c|c|c|c|c|c}
\multicolumn{2}{c|}{Datasets} & Methods & {Features}& {Global} & {Covis.}& {Local}& {\ \ PnP\ \ }& {\textbf{Total}}\\ \hline

\multirow{8}{*}{\rotatebox[origin=c]{90}{Aachen}}
& \multirow{4}{*}{Day}
& AS & 263 & - & - & \hcellrule{2}{112} & \textbf{375} \\
&& NV+SIFT & 92+263 & 7 & 8 & 1220 & 29 & \textbf{1356} \\ 
&& \textbf{NV+SP} & 92+26 & 7 & 5 & 9 & 9 & \textbf{148} \\ 
&& \textbf{HF-Net} & 15 & 7 & 5 & 9 & 9 & \hlf{45} \\
\cline{2-9}

& \multirow{4}{*}{Night}
& AS & 263 & - & - & \hcellrule{2}{132} & \textbf{395} \\
&& NV+SIFT & 92+263 & 7 & 8 & 1492 & 56 & \textbf{1655} \\ 
&& \textbf{NV+SP} & 92+26 & 7 & 5 & 10 & 18 & \textbf{158} \\
&& \textbf{HF-Net} & 15 & 7 & 5 & 10 & 18 & \hlf{55} \\ 
\hline

\multirow{8}{*}{\rotatebox[origin=c]{90}{RobotCar}}
& \multirow{4}{*}{Dusk}
& AS & 189 & - & - & \hcellrule{2}{283} & \textbf{472} \\
&& NV+SIFT & 92+189 & 13 & 3 & 264 & 14 & \textbf{575} \\ 
&& \textbf{NV+SP} & 92+26 & 13 & 1 & 3 & 4 & \textbf{139} \\
&& \textbf{HF-Net} & 15 & 13 & 1 & 3 & 4 & \hlf{36} \\
\cline{2-9}

& \multirow{4}{*}{Night}
& AS & 189 & - & - & \hcellrule{2}{1021} & \textbf{1210} \\
&& NV+SIFT & 92+189 & 13 & 3 & 389 & 149 & \textbf{835} \\ 
&& \textbf{NV+SP} & 92+26 & 13 & 1 & 6 & 38 & \textbf{176} \\
&& \textbf{HF-Net} & 15 & 13 & 1 & 6 & 38 & \hlf{73} \\
\end{tabular}}%
}%
\end{center}
\vspace{-6pt}
\caption{\textbf{Timings [ms]} for the different steps of hierarchical localization: feature extraction, global search, covisibility clustering, local matching, and pose estimation with PnP. Feature extraction with SIFT or CNN and matching of learned descriptors are performed on the GPU, and other operations on the CPU. We highlight the \hlf{fastest} method for each sequence. Localizing with HF-Net is 10 times faster than with AS, the fastest method available.
}
\label{tab:timings}
\end{table}

\PAR{Hierarchical localization.} Timings of NV+SP and HF-Net show that our coarse-to-fine approach scales well to large environments. The global search is fast, and only depends on the number of images used to build the model. It successfully reduces the set of potential candidate correspondences and enables a tractable 2D-3D matching. This strongly depends on the SfM model -- the denser the covisibility graph is, the more 3D points are retrieved and matched per prior frame, which increases the runtime. NV+SIFT is therefore prohibitively slow, as its SfM model is much denser, especially on Aachen. NV+SP significantly improves on it, as the sparser SfM model yield clusters with fewer 3D points. The inference of NetVLAD and SuperPoint however accounts for 75\% of its runtime, and is thus, as previously mentioned, the bottleneck. HF-Net mitigates this issues with an inference 7 times faster.

\PAR{Existing approaches.} CSL and SMC are not listed in Table~\ref{tab:timings} as they both require several tens of seconds per query, and are thus three orders of magnitude slower than our fastest method. AS improves on this, but is still slower, especially in case of a low success rate, such as on RobotCar night. Overall, our localization system based on HF-Net can run at 20 FPS on very large-scale environments. It is 10 times faster than AS, designed for efficiency, and is much more accurate on all datasets.

\section{Conclusion}
In this paper, we have presented a method for visual localization that is at the same time robust, accurate, and runs in real-time.
Our system follows a coarse-to-fine localization paradigm. First, it performs a global image retrieval to obtain a set of database images, which are subsequently clustered into places using the covisibility graph of a 3D SfM model. We then perform local 2D-3D matching within the candidate places to obtain an accurate 6-DoF estimate of the camera pose.

A version of our method is based on existing neural networks for image retrieval and feature matching. It outperforms state-of-the-art localization approaches on several large-scale benchmarks that include day-night queries and substantial appearance variations across weather conditions and seasons. We then improve its efficiency by proposing HF-Net, a novel CNN that computes keypoints as well as global and local descriptors in a single shot. We demonstrate the effectiveness of multitask distillation to train it in a flexible manner while retaining the original performance. The resulting localization systems runs at more than 20 FPS at large scale and offers an unparalleled robustness in challenging conditions.

{\small
\vskip4pt \noindent{\textbf{Acknowledgements}.\ We thank the reviewers for their valuable comments, Torsten Sattler for helping to evaluate the localization, and Eduard Trulls for providing support for the SfM dataset.}
}

\newpage
\appendix
\section*{Appendix}
We provide here additional experiment details and qualitative results.

\section{HF-Net Implementation}
\subsection{Network Architecture}
\begin{figure}[!htb]
\begin{center}
    \includegraphics[width=1.0\linewidth]{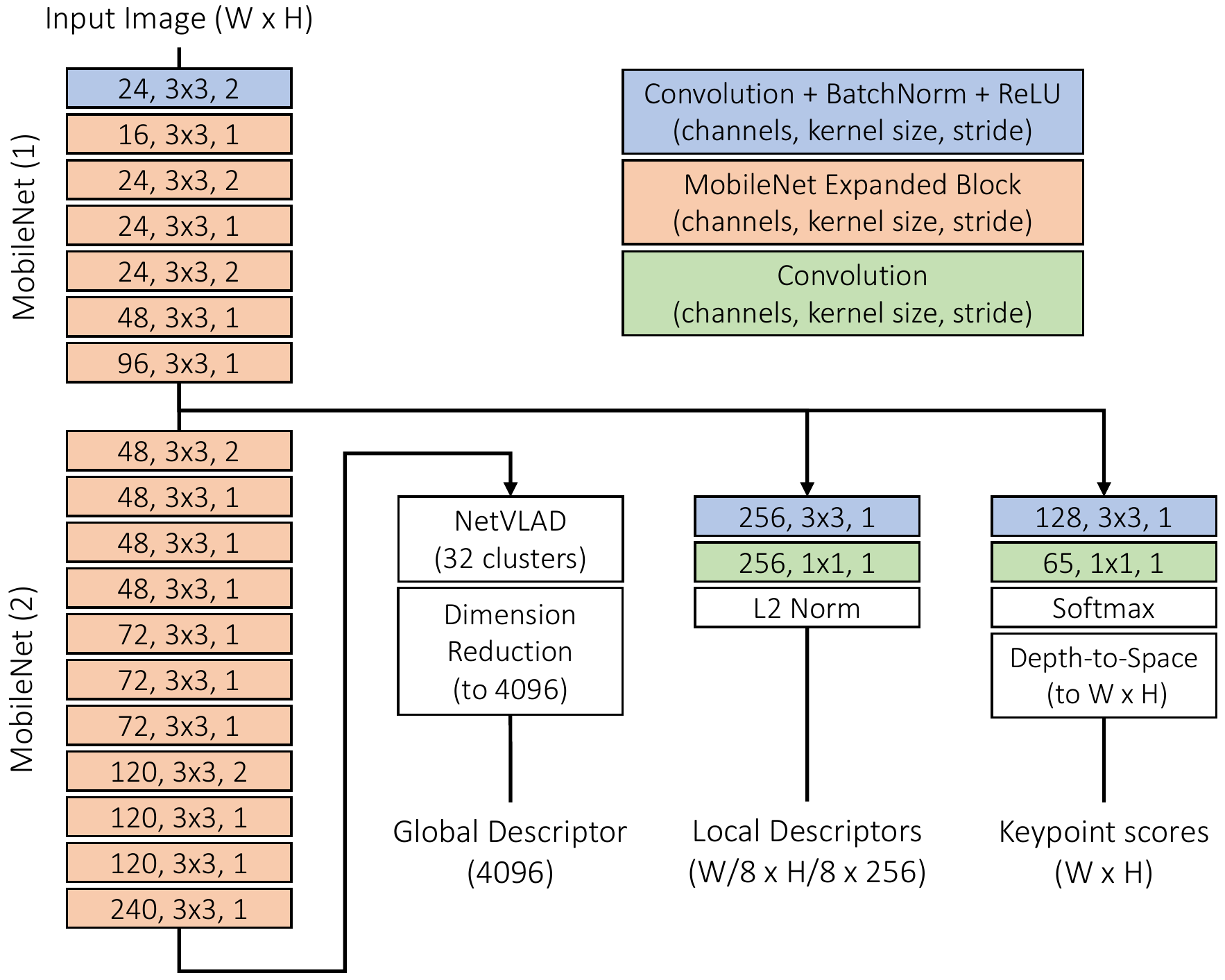}
\end{center}
   \caption{\textbf{Detail of the HF-Net architecture}, consisting of a MobiletNet encoder and three heads predicting a global descriptor, a dense local descriptor map, and keypoint scores.}
\label{fig:hfnet-detail}
\vspace{-4mm}
\end{figure}

HF-Net is built on top of a MobileNetV2~\cite{mobilenet} encoder with depth multiplier 0.75. The local heads are identical to the original SuperPoint~\cite{superpoint} and branch off at the layer 7. The global head is composed of a NetVLAD layer~\cite{netvlad} and a dimensionality reduction, implemented as a multiplication with a learnable matrix, in order to match the dimension of the target teacher descriptor. The global head is appended to the MobileNet layer 18. The detailed architecture is shown in Figure~\ref{fig:hfnet-detail}.

\subsection{Training details}
The images from both Google Landmarks~\cite{delf} and Berkeley Deep Drive~\cite{bdd} are resized to $640 \times 480$ and converted to grayscale. We found RGB to be detrimental to the performance of the local feature heads, most likely because of the limited bandwidth of the encoder. As photometric data augmentation, we apply, in a random order, Gaussian noise, motion blur in random directions, and random brightness and contrast changes.

The losses of the global and local descriptors are the L2 distances with their targets. For the keypoints, we apply the cross-entropy with the target probabilities (soft labels). We found hard labels to perform poorly, likely due to their sparsity and the smaller size of the student network. The three losses are aggregated using the multi-task learning scheme of~Kendall \etal~\cite{kendallmulti}.

The MobileNet layers are initialized with weights pre-trained on ImageNet~\cite{imagenet}. The network is implemented with Tensorflow~\cite{abadi2016tensorflow} and trained for 85k iterations with the RMSProp optimizer~\cite{tieleman2012lecture} and a batch size of 32. We use an initial learning rate of $10^{-3}$, which is successively divided by ten at iterations 60k and 80k.

\section{Local Feature Evaluation}
\subsection{Setup}
The images of both HPatches~\cite{balntas20017hpatches} and SfM~\cite{ono2018lf} datasets are resized so that their largest dimension is 640 pixels. The metrics are computed on image pairs and follow the definitions of~\cite{superpoint, ono2018lf}. A keypoint $k_1$ in an image is deemed correct if its reprojection $\hat{k}_1$ in a second image lies within a given distance threshold $\epsilon$ to a second detected keypoint $k_2$. Additionally, $k_1$ is matched correctly if it is correct and if $k_2$ is its nearest neighbor in the descriptor space.

For HPatches, we detect 300 keypoints for both keypoint and descriptor evaluations, and set $\epsilon = 3$ pixels. The homography is estimated using the OpenCV function \verb|findHomography| and considered accurate if the average reprojection error of the image corners is lower than 3 pixels. For the SfM dataset, due to the extensive texture, 1000 keypoints are detected. The keypoint and descriptor metrics use correctness thresholds $\epsilon$ of 3 and 5, respectively. The 6-DoF pose is estimated with the function \verb|solvePnPRansac|, and deemed correct if its ground truth is within distance and orientation thresholds of 3 m and 1\degree, respectively.

For DoG, Harris~\cite{harris1988combined}, and SIFT~\cite{lowe2004distinctive}, we use the implementations of OpenCV. For SuperPoint~\cite{superpoint} and LF-Net~\cite{ono2018lf}, we use the implementations provided by the authors. For NetVLAD, we use the implementation of~\cite{netvlad-tf} and the original model trained on Pittsburgh30k. Dense descriptors are obtained by normalizing the feature map \verb|conv3_3| before the ReLU activation. For DOAP~\cite{doap}, we use the trained model provided by the authors. As we are mostly interested in dense descriptors for run-time efficiency, we disable the spatial transformer and enable padding in the last layer, thus producing a feature map four times smaller than the input image. We found the model trained on HPatches with spatial transformer to give the best results and thus only evaluate DOAP on the SfM dataset. As a post-processing, we apply Non-Maximum Suppression (NMS) with a radius of 4 to both Harris and SuperPoint. Sparse descriptors are sampled from the dense maps of SuperPoint, NetVLAD, and DOAP using bilinear interpolation.

\subsection{Qualitative Results}
\begin{figure*}[htb!]
\centering
\def\iwidth{.32\linewidth}

\begin{minipage}{\iwidth}
    \centering
    \includegraphics[width=0.98\linewidth]{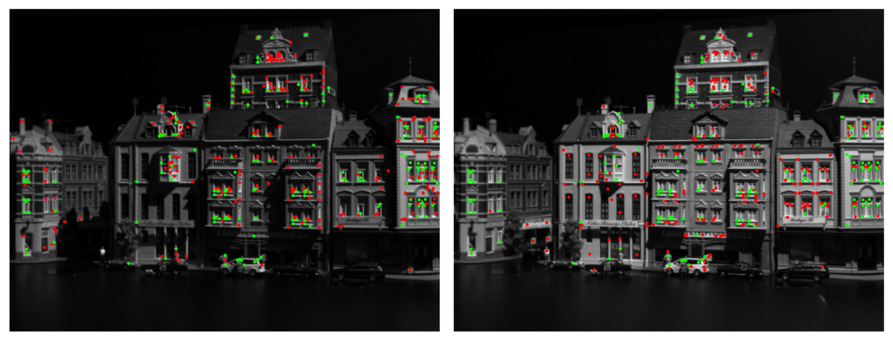}
\end{minipage}%
\begin{minipage}{\iwidth}
    \centering
    \includegraphics[width=0.98\linewidth]{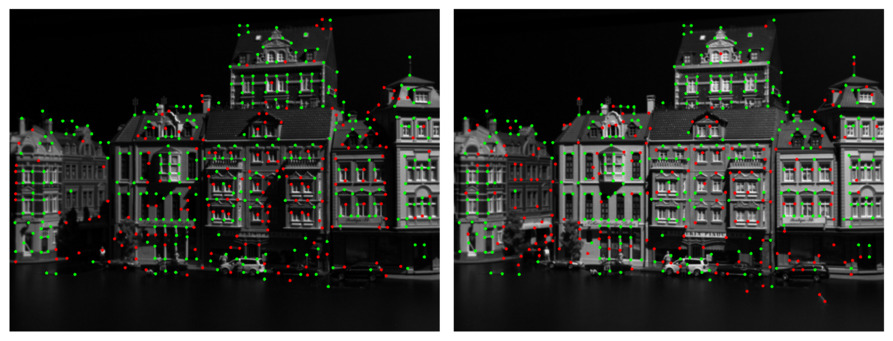}
\end{minipage}%
\begin{minipage}{\iwidth}
    \centering
    \includegraphics[width=0.98\linewidth]{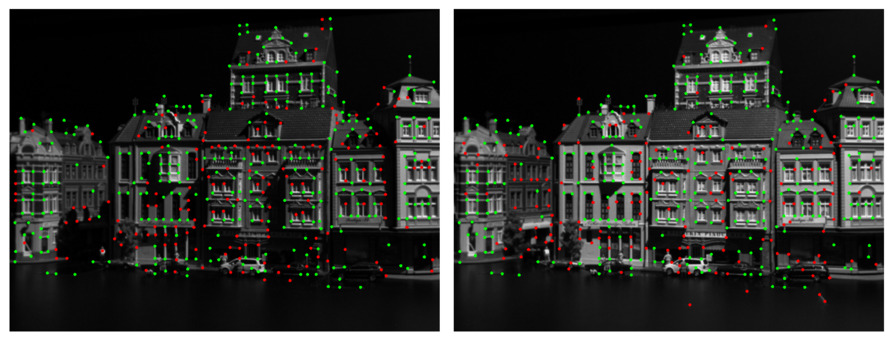}
\end{minipage}

\begin{minipage}{\iwidth}
    \centering
    \includegraphics[width=0.98\linewidth,trim={1.5mm 0 1.5mm 0},clip]{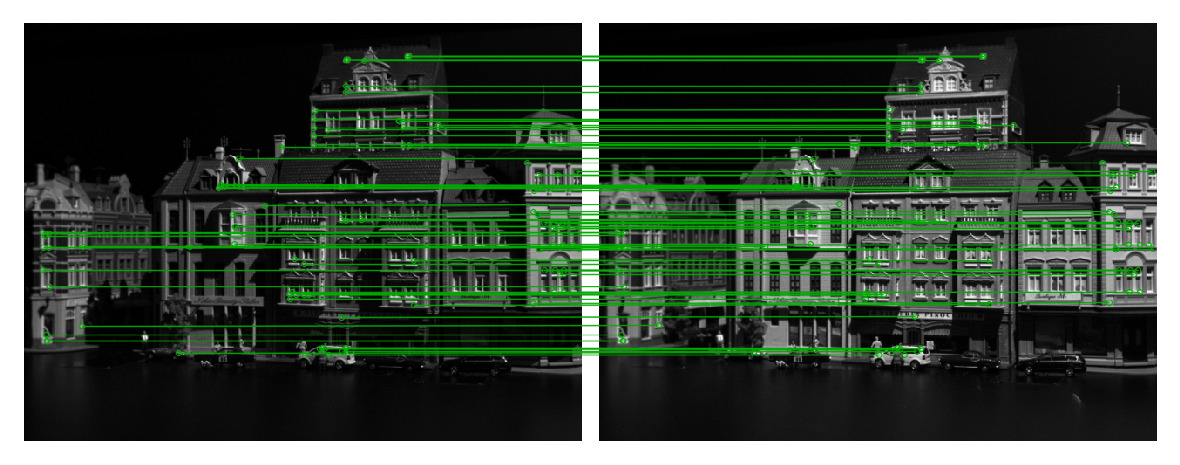}
\end{minipage}%
\begin{minipage}{\iwidth}
    \centering
    \includegraphics[width=0.98\linewidth,trim={1.5mm 0 1.5mm 0},clip]{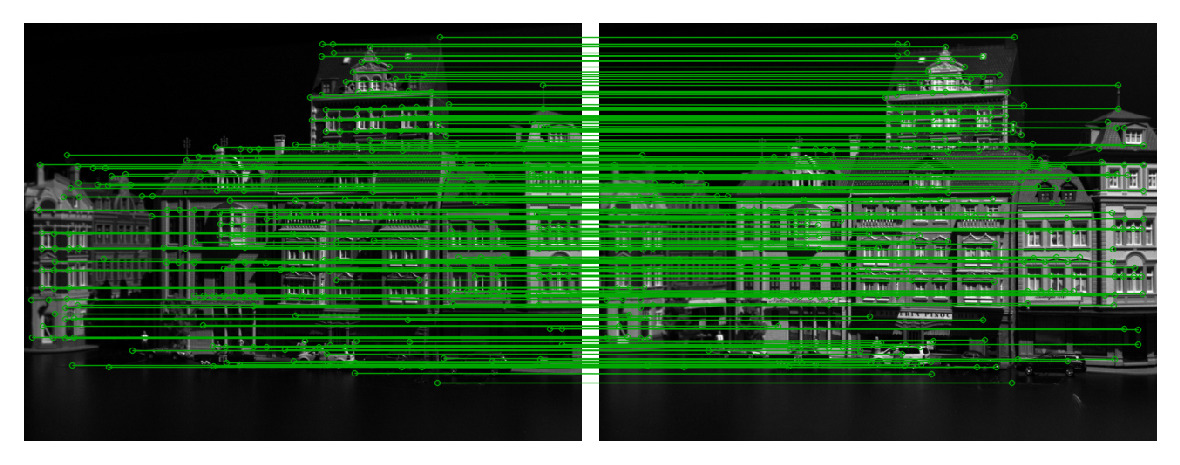}
\end{minipage}%
\begin{minipage}{\iwidth}
    \centering
    \includegraphics[width=0.98\linewidth,trim={1.5mm 0 1.5mm 0},clip]{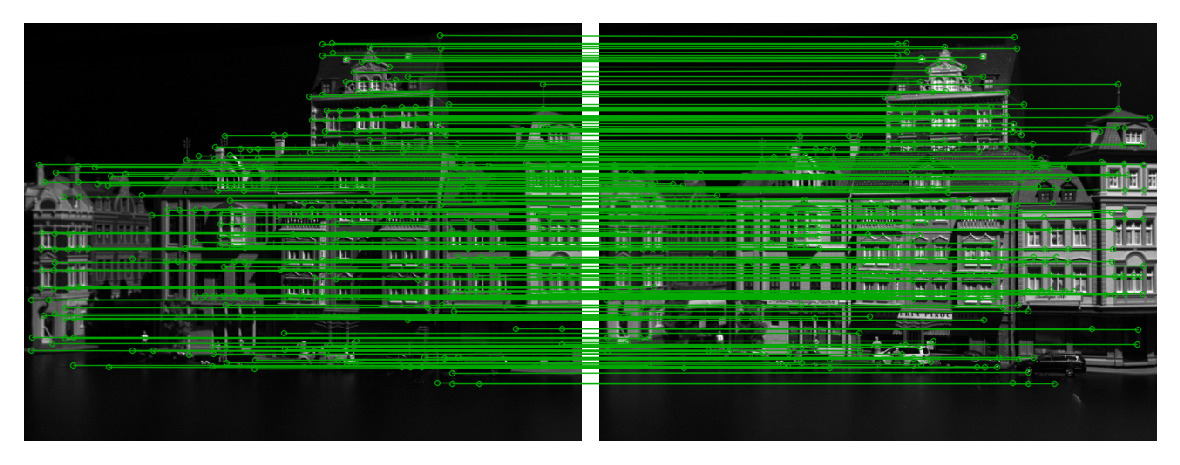}
\end{minipage}

\begin{minipage}{\iwidth}
    \centering
    \includegraphics[width=0.98\linewidth]{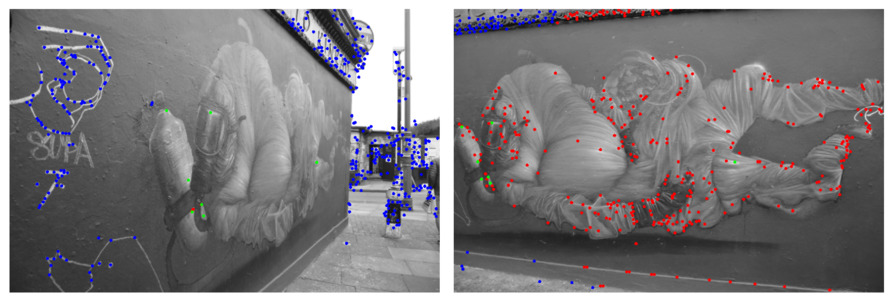}
\end{minipage}%
\begin{minipage}{\iwidth}
    \centering
    \includegraphics[width=0.98\linewidth]{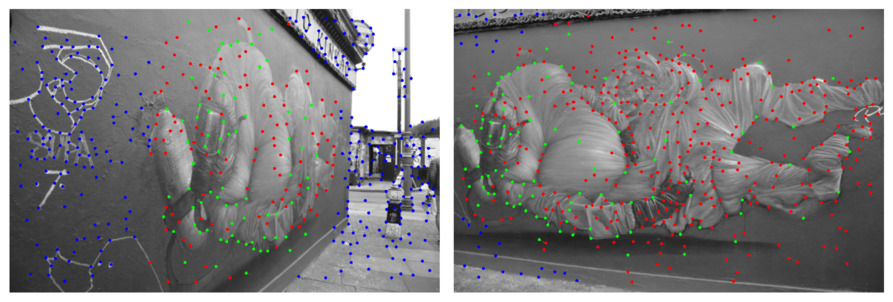}
\end{minipage}%
\begin{minipage}{\iwidth}
    \centering
    \includegraphics[width=0.98\linewidth]{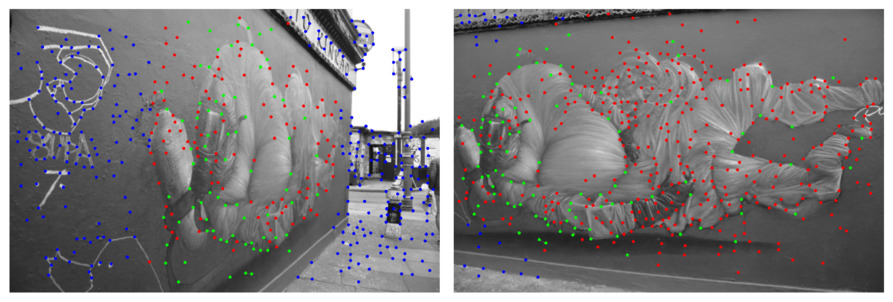}
\end{minipage}

\begin{minipage}{\iwidth}
    \centering
    \includegraphics[width=0.98\linewidth,trim={1.5mm 0 1.5mm 0},clip]{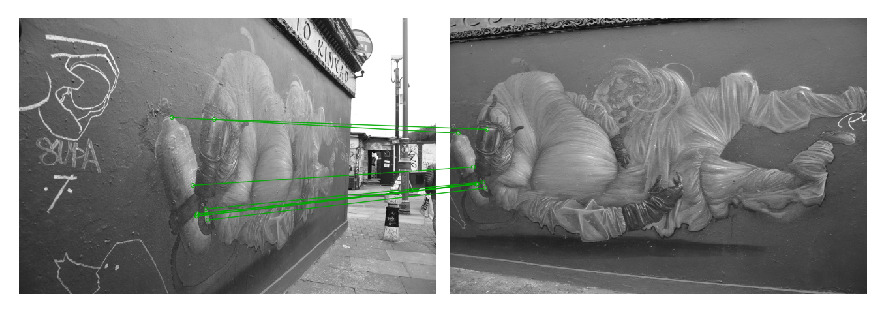}
\end{minipage}%
\begin{minipage}{\iwidth}
    \centering
    \includegraphics[width=0.98\linewidth,trim={1.5mm 0 1.5mm 0},clip]{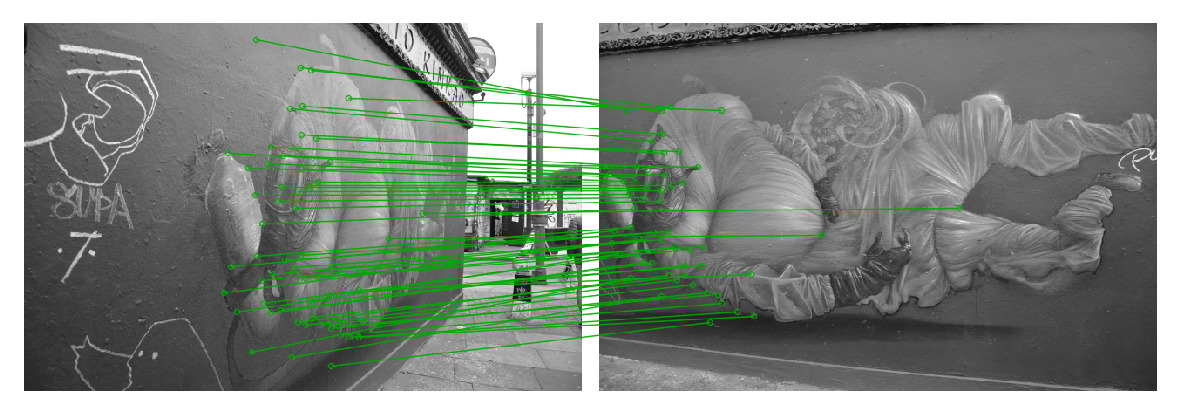}
\end{minipage}%
\begin{minipage}{\iwidth}
    \centering
    \includegraphics[width=0.98\linewidth,trim={1.5mm 0 1.5mm 0},clip]{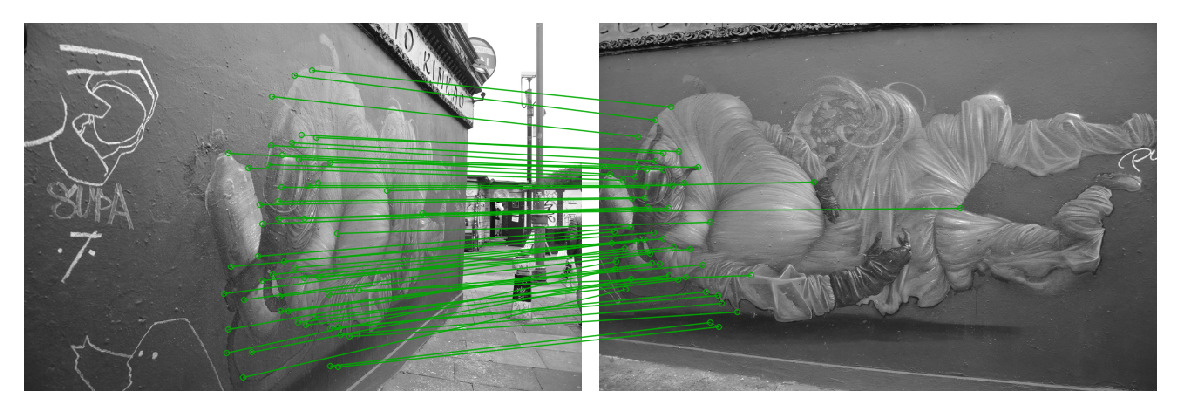}
\end{minipage}

\caption{\textbf{Qualitative results on the HPatches dataset.} Keypoints (green if repeatable, red if not repeatable,  blue if not visible in the other image) and inlier matches are shown for SIFT (left), SuperPoint (center) and HF-Net (right).}
\label{fig:qual:hpatches}
\end{figure*}

\begin{figure*}[htb!]
\centering
\def\iwidth{.32\linewidth}

\begin{minipage}{\iwidth}
    \centering
    \includegraphics[width=0.98\linewidth]{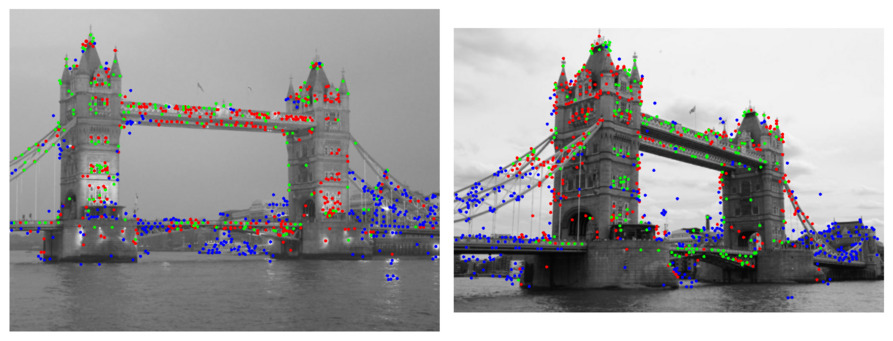}
\end{minipage}%
\begin{minipage}{\iwidth}
    \centering
    \includegraphics[width=0.98\linewidth]{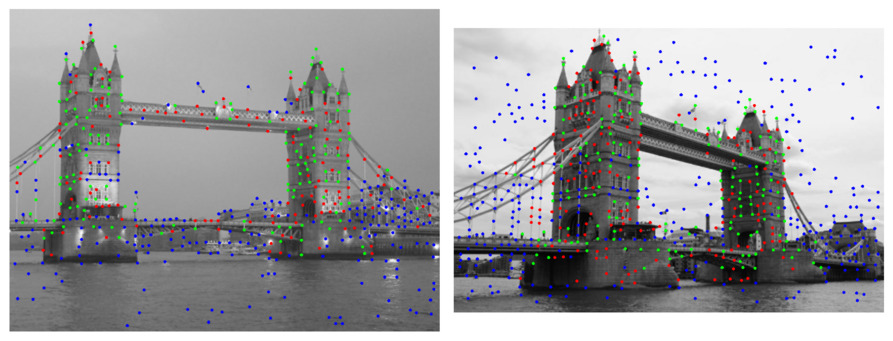}
\end{minipage}%
\begin{minipage}{\iwidth}
    \centering
    \includegraphics[width=0.98\linewidth]{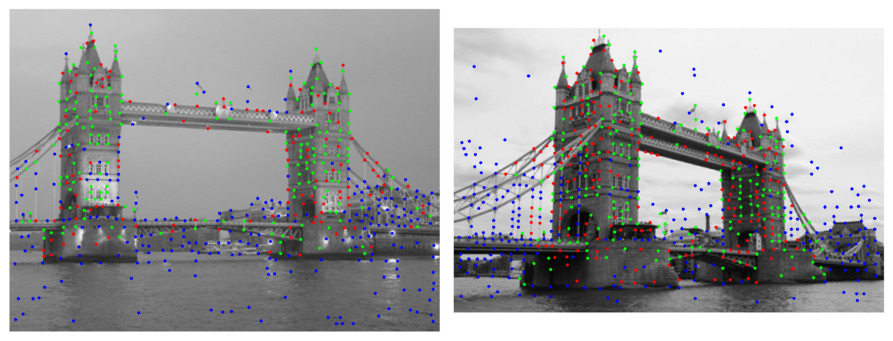}
\end{minipage}

\begin{minipage}{\iwidth}
    \centering
    \includegraphics[width=0.98\linewidth,trim={1.5mm 0 1.5mm 0},clip]{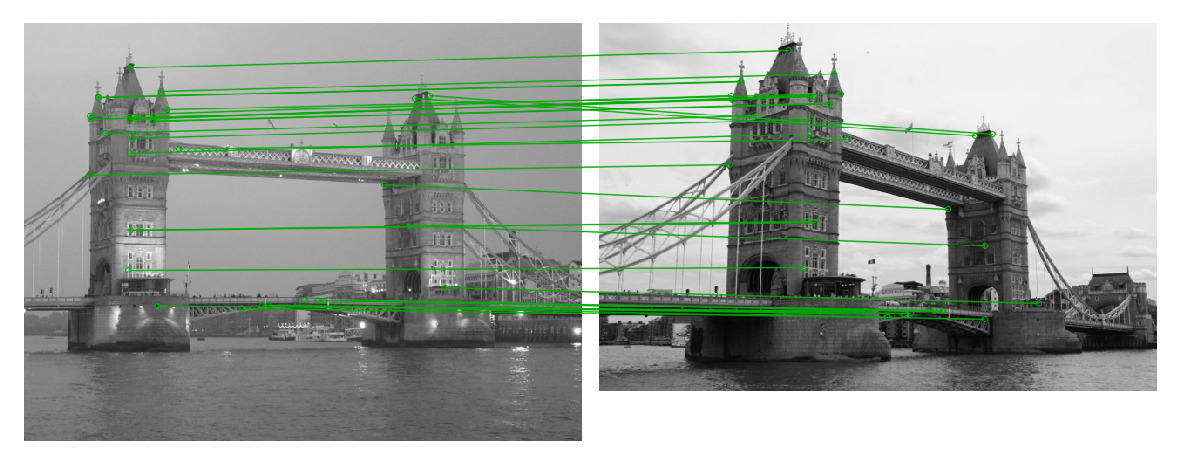}
\end{minipage}%
\begin{minipage}{\iwidth}
    \centering
    \includegraphics[width=0.98\linewidth,trim={1.5mm 0 1.5mm 0},clip]{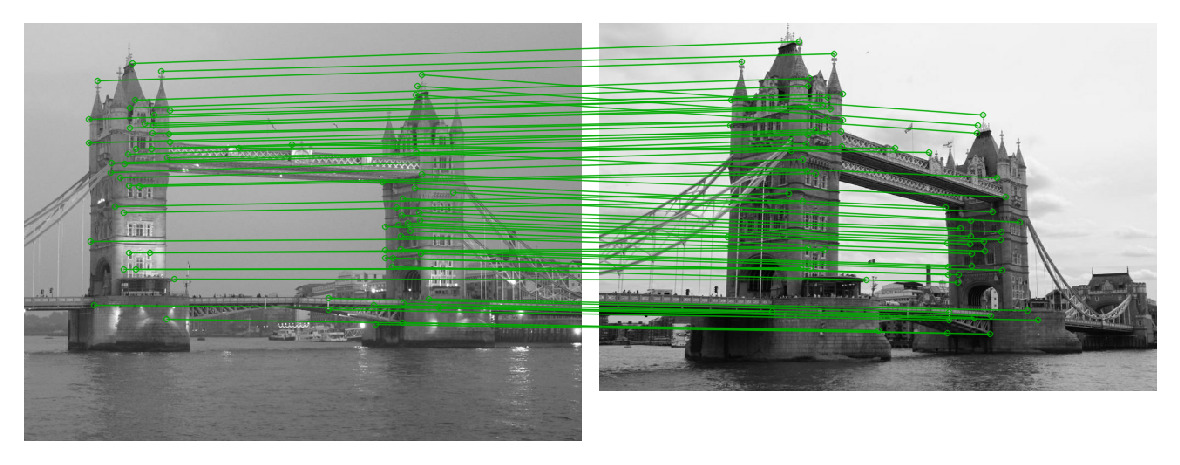}
\end{minipage}%
\begin{minipage}{\iwidth}
    \centering
    \includegraphics[width=0.98\linewidth,trim={1.5mm 0 1.5mm 0},clip]{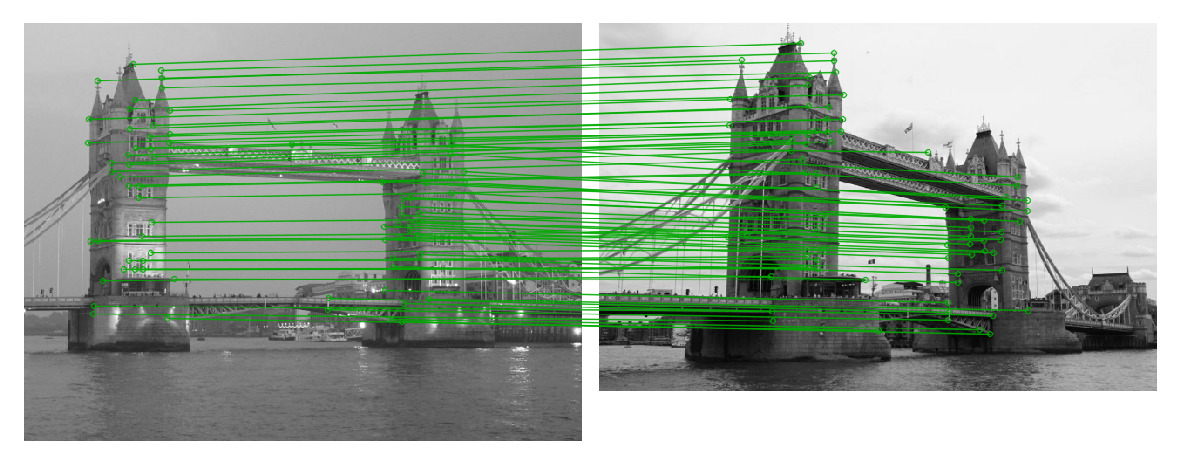}
\end{minipage}

\begin{minipage}{\iwidth}
    \centering
    \includegraphics[width=0.98\linewidth]{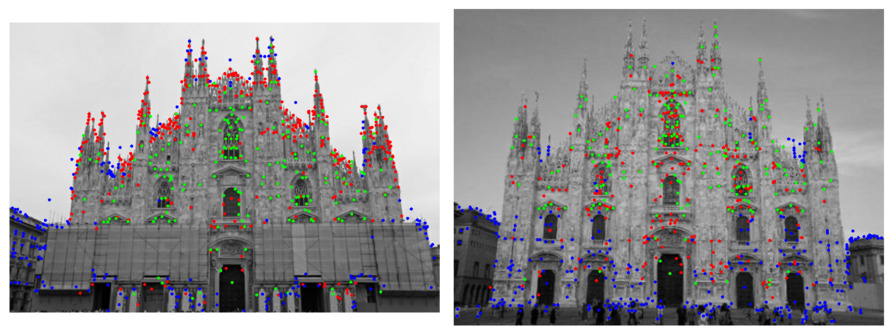}
\end{minipage}%
\begin{minipage}{\iwidth}
    \centering
    \includegraphics[width=0.98\linewidth]{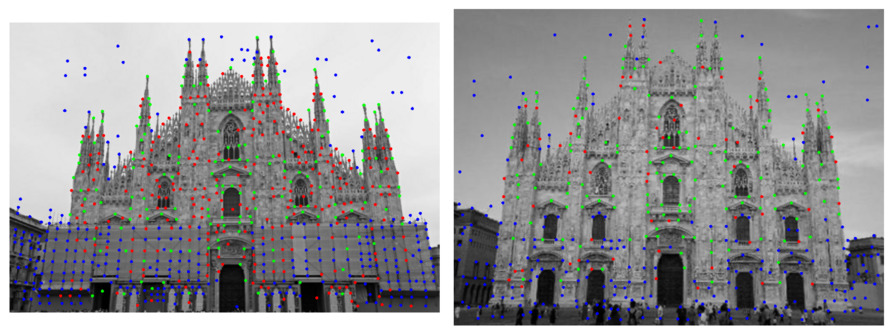}
\end{minipage}%
\begin{minipage}{\iwidth}
    \centering
    \includegraphics[width=0.98\linewidth]{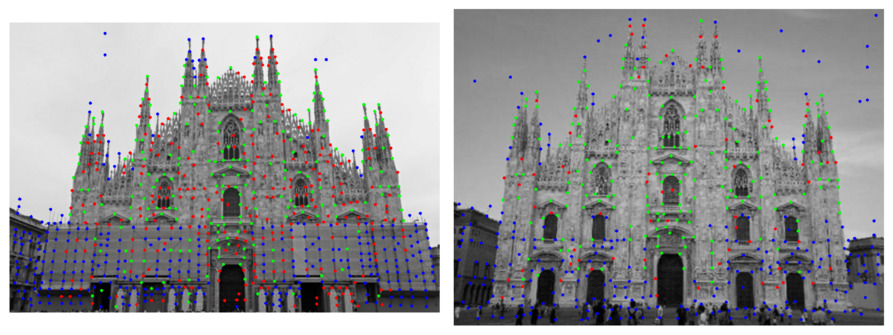}
\end{minipage}

\begin{minipage}{\iwidth}
    \centering
    \includegraphics[width=0.98\linewidth,trim={1.5mm 0 1.5mm 0},clip]{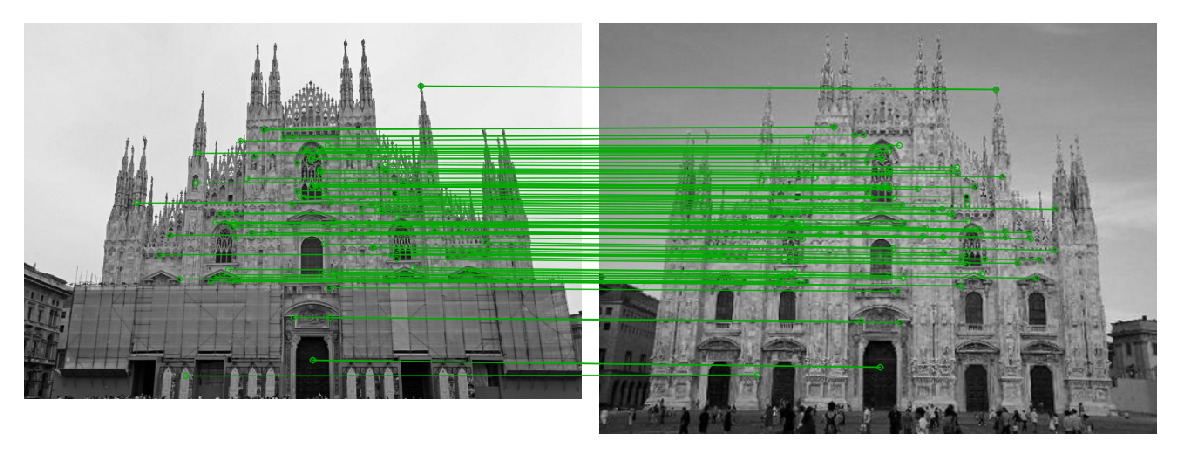}
\end{minipage}%
\begin{minipage}{\iwidth}
    \centering
    \includegraphics[width=0.98\linewidth,trim={1.5mm 0 1.5mm 0},clip]{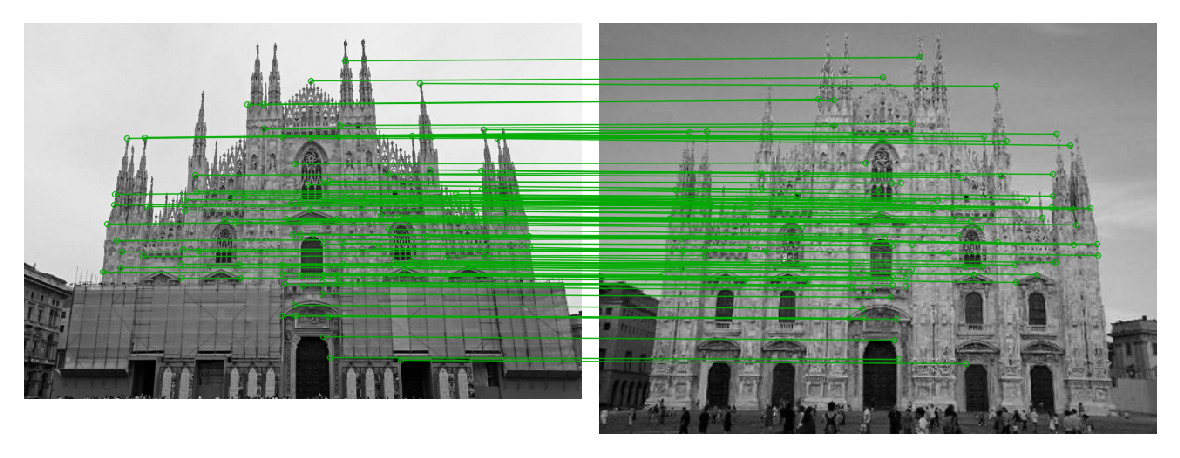}
\end{minipage}%
\begin{minipage}{\iwidth}
    \centering
    \includegraphics[width=0.98\linewidth,trim={1.5mm 0 1.5mm 0},clip]{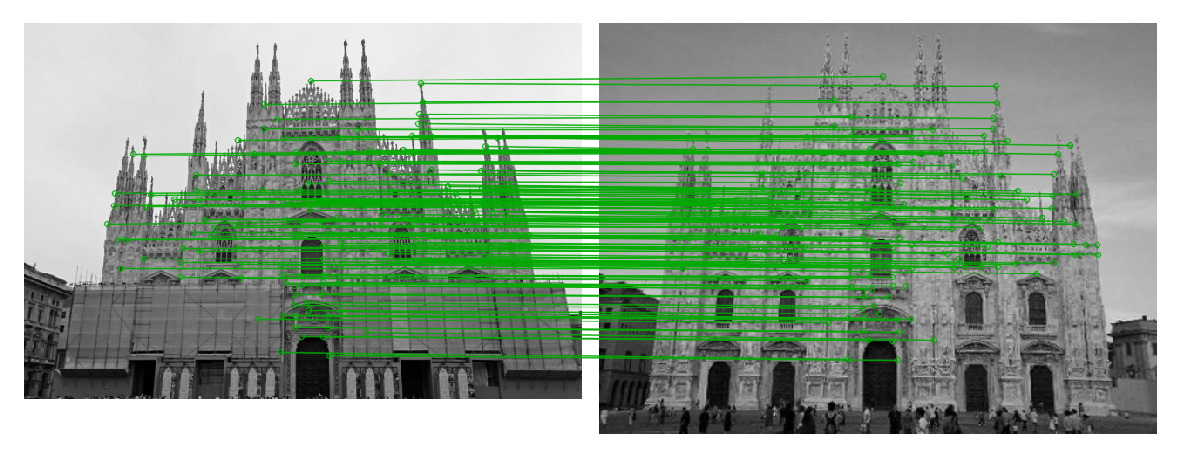}
\end{minipage}

\caption{\textbf{Qualitative results on the SfM dataset} for SIFT (left), SuperPoint (center) and HF-Net (right).}
\label{fig:qual:sfm}
\end{figure*}

We show in Figures~\ref{fig:qual:hpatches} and~\ref{fig:qual:sfm} detected keypoints and their corresponding matches on the HPatches and SfM datasets, respectively.

\section{Large-scale Localization}
\subsection{Model Quality}
Extended statistics of models built with SIFT and HF-Net for the Aachen Day-Night, RobotCar Seasons, and CMU Seasons datasets, are provided in Table~\ref{tab:models}. We also report the track length, \ie the number of observation per 3D point, as defined by~\cite{schonberger2017comparative}. The metrics for the CMU dataset are aggregated over the models of the slices corresponding to the urban and suburban environments. For SIFT, some metrics cannot be computed on the CMU model as the keypoints that are not matched were not provided.

\begin{table}[htb!]
\centering
\resizebox{\columnwidth}{!}{%
\scriptsize{\setlength\tabcolsep{3.0pt}
\begin{tabular}{l|cc|cc|cc}
\toprule
& \multicolumn{2}{c|}{\textbf{Aachen}}& \multicolumn{2}{c|}{\textbf{RobotCar}}&
\multicolumn{2}{c}{\textbf{CMU}}\\
Statistics & {SIFT}& {HF}& {SIFT}& {HF} & {SIFT} & {HF}\\
\midrule
{\# 3D points}     & 1,900k & 685k & 6,869k & 2,525k & 961k & 553k\\
{\# Keypoints per image} & 10,230 & 2,576 & 4,409 & 970 & - & 1,446\\
{Ratio of matched keypoints [\%]}     & 18.8 & 33.8 & 39.4 & 59.9 & - & 45.3\\
{Track length} & 5.85 & 5.87 & 5.34 & 4.71 & 4.11 & 4.95\\
\bottomrule
\end{tabular}}%
}%
\vspace{2mm}
\caption{\textbf{Statistics of 3D models} built with SIFT and HF-Net.}
\label{tab:models}
\end{table}

\subsection{Implementation Details}
We now provide additional details regarding the implementation of our hierarchical localization pipeline. For all datasets, we reduce the dimensionality of the global descriptors predicted by both NetVLAD and HF-Net to 1024 dimensions using PCA, whose parameters are learned on the reference images, independently for each dataset. A total of 10 prior frames are retrieved and clustered. Due to limits on the GPU memory, features are extracted on images downsampled such that their largest dimension is 960 pixels for Aachen and Robotcar, and 1024 for CMU. For both SuperPoint and HF-Net, NMS with radius 4 is applied to the detected keypoints in the query image and 2k of them are retained. When performing local matching, our modified ratio test uses a threshold of 0.9. PnP+RANSAC uses a threshold on the reprojection error of 10 pixels for Aachen, 5 pixels for CMU (due to the lower image size), and 12 pixels for RobotCar (due to the lower keypoint localization accuracy of SuperPoint and HF-Net). The estimated pose is deemed correct when the number of inliers is larger than a threshold, whose value is 12 for Aachen and CMU, and 15 for Robotcar.

\subsection{Evaluation Process}
The method and baselines introduced in this work are evaluated on all three datasets by the benchmark's authors~\cite{loc-benchmark}, who also generated the plots shown in the main paper. For Active Search~\cite{active-search}, City Scale Localization~\cite{city-scale-loc}, DenseVLAD~\cite{torii201524}, and NetVLAD~\cite{netvlad}, we use the evaluation reported in the paper introducing the benchmark.

The evaluation of Semantic Match Consistency~\cite{toft2018semantic} (SMC) is the one reported in the original paper. We do not directly compare this method to the ones introduced in the present work, nor to the benchmark baselines, as SMC assumes a known camera height, and, more importantly, relies on a semantic segmentation CNN which was trained on the evaluation dataset of RobotCar. We emphasize that our HF-Net never encountered any test data during training, and that it was evaluated on the three datasets using the same trained model.

\subsection{Qualitative Results}
\begin{figure*}[htb!]
\centering
\def\iwidth{.32\linewidth}

\begin{minipage}{\iwidth}
    \centering
    \includegraphics[width=0.98\linewidth]{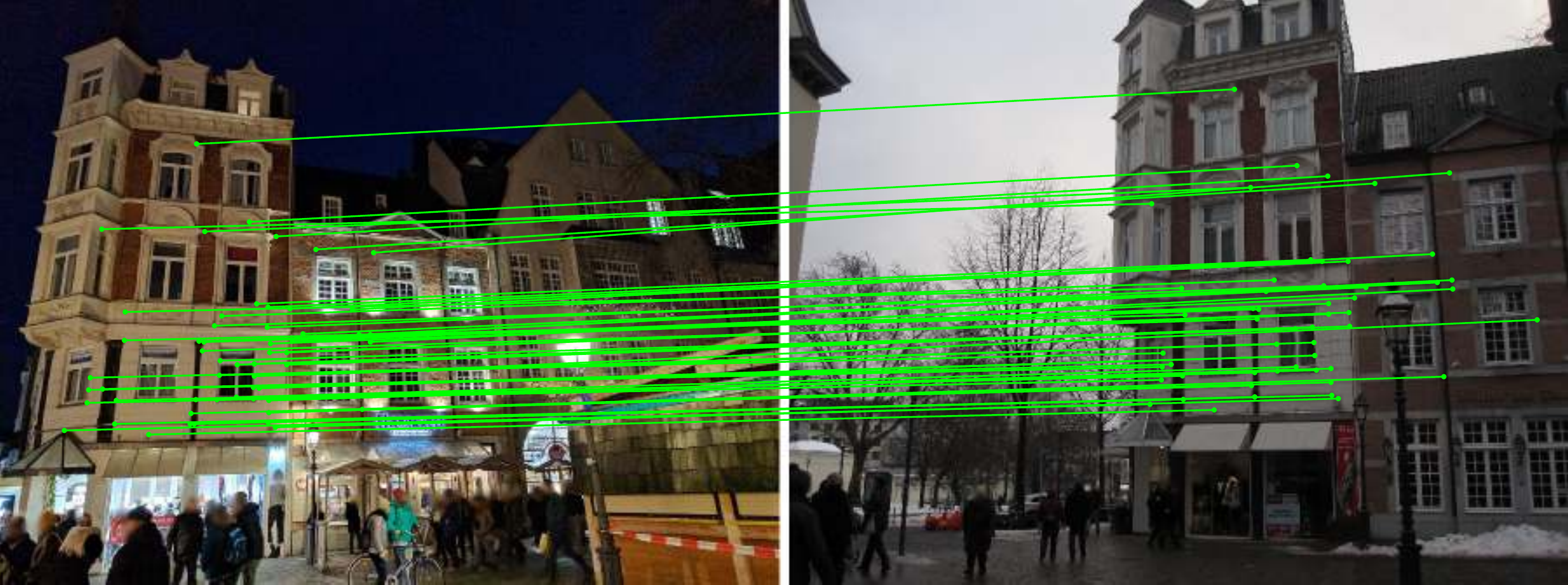}
\end{minipage}%
\begin{minipage}{\iwidth}
    \centering
    \includegraphics[width=0.98\linewidth]{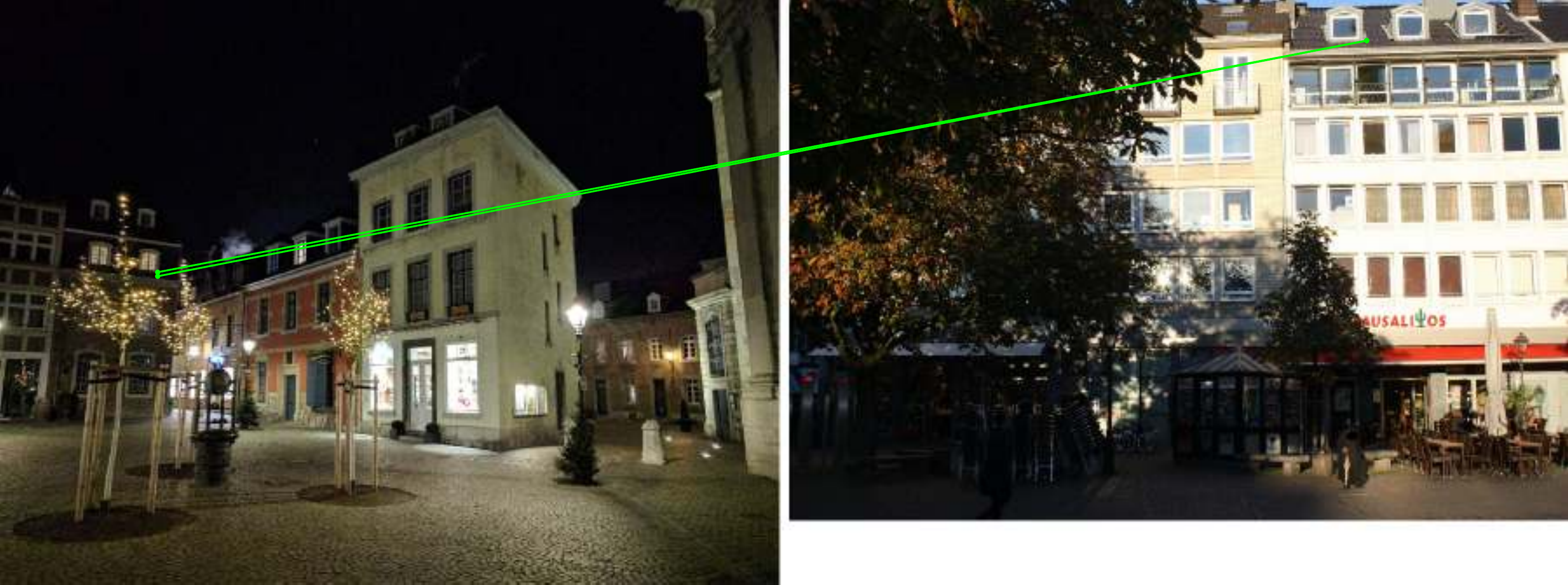}
\end{minipage}%
\begin{minipage}{\iwidth}
    \centering
    \includegraphics[trim=0 100 0 0,clip,width=0.98\linewidth]{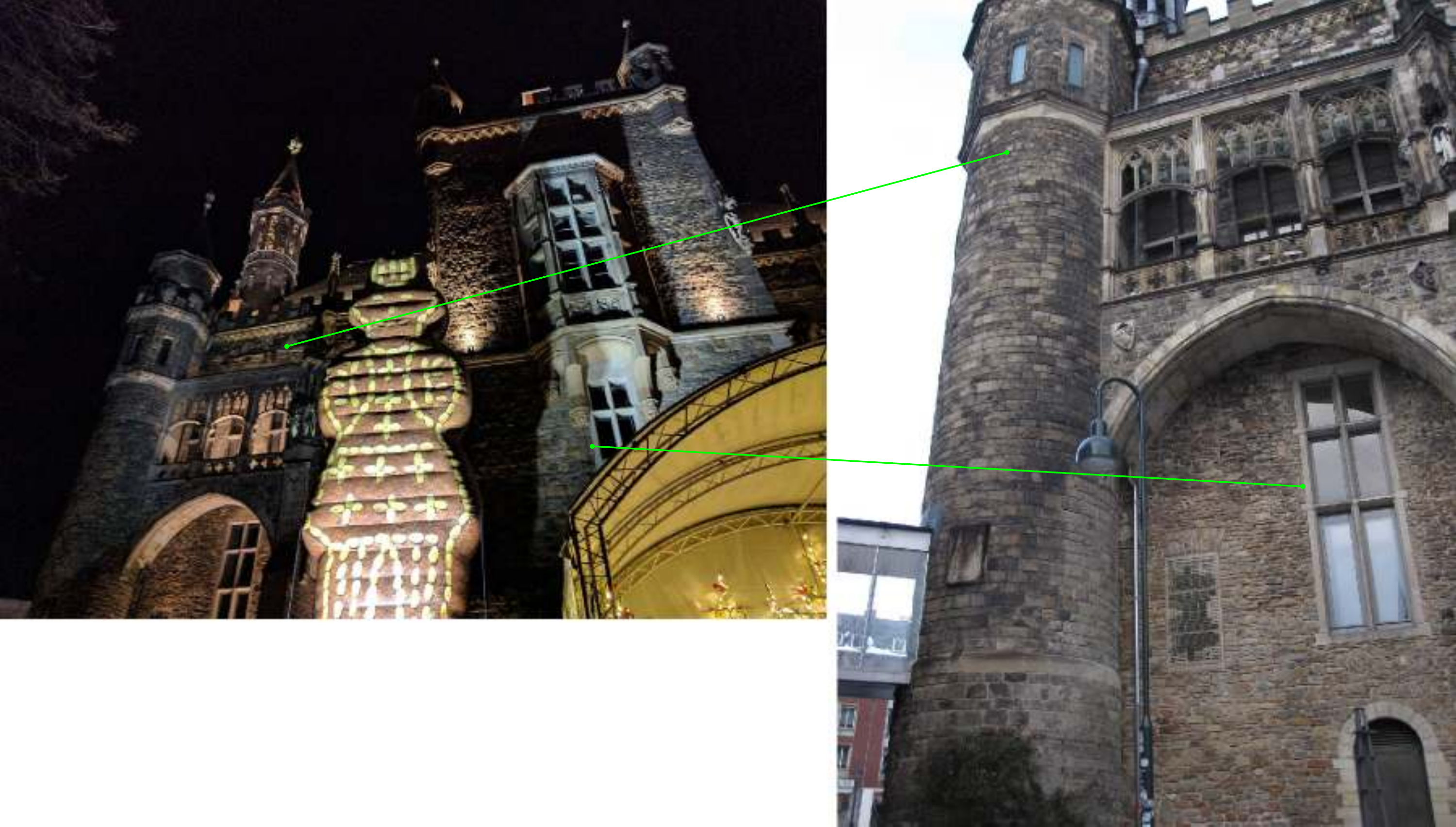}
\end{minipage}%
\vspace{3mm}

\begin{minipage}{\iwidth}
    \centering
    \includegraphics[width=0.98\linewidth]{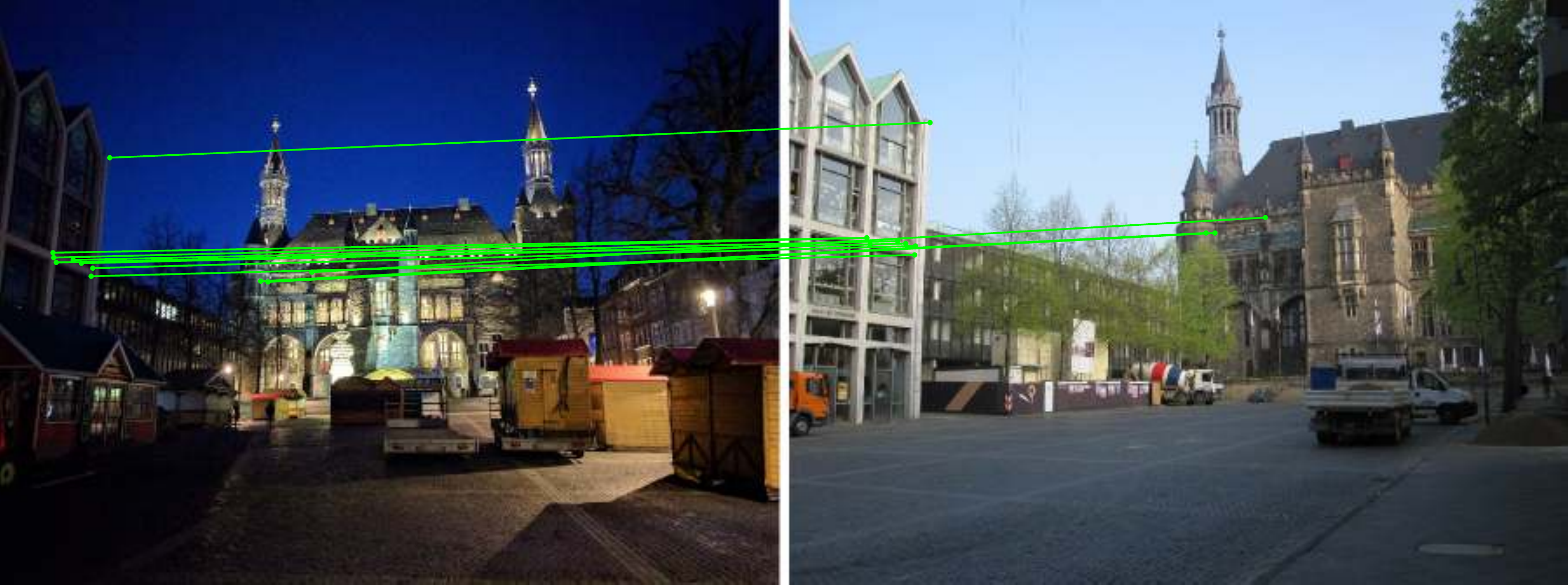}
\end{minipage}%
\begin{minipage}{\iwidth}
    \centering
    \includegraphics[width=0.98\linewidth]{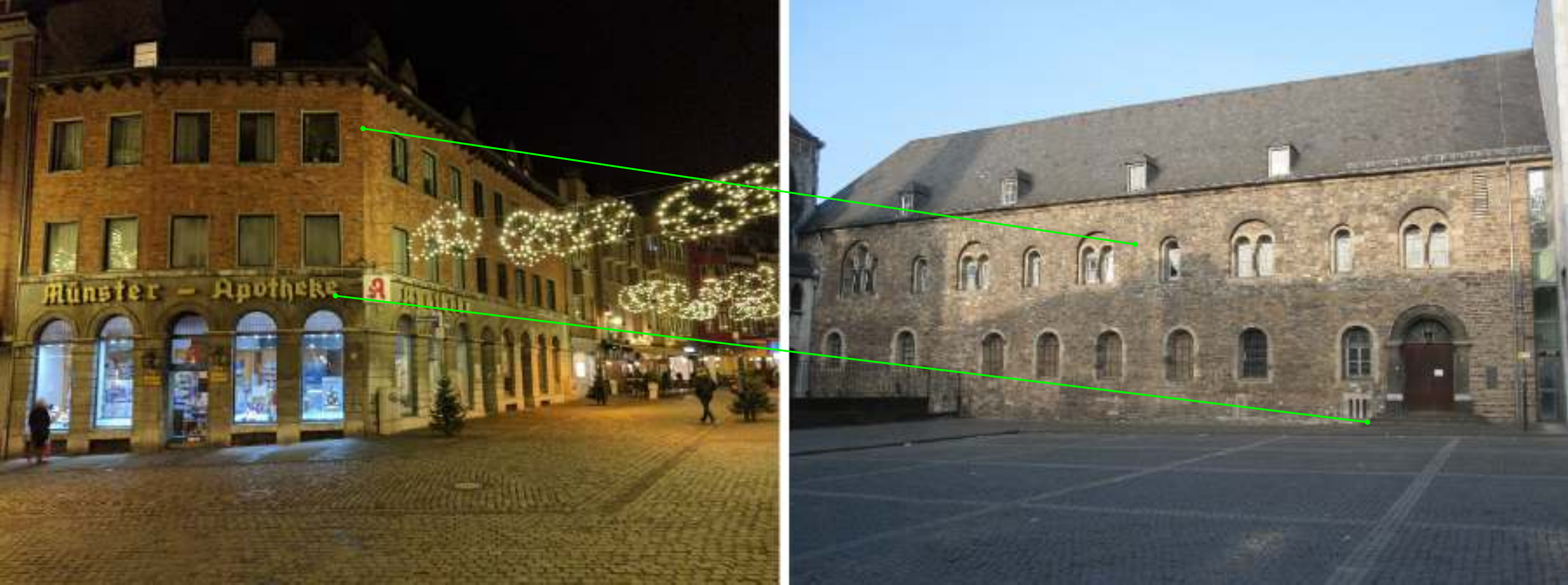}
\end{minipage}%
\begin{minipage}{\iwidth}
    \centering
    \includegraphics[width=0.98\linewidth]{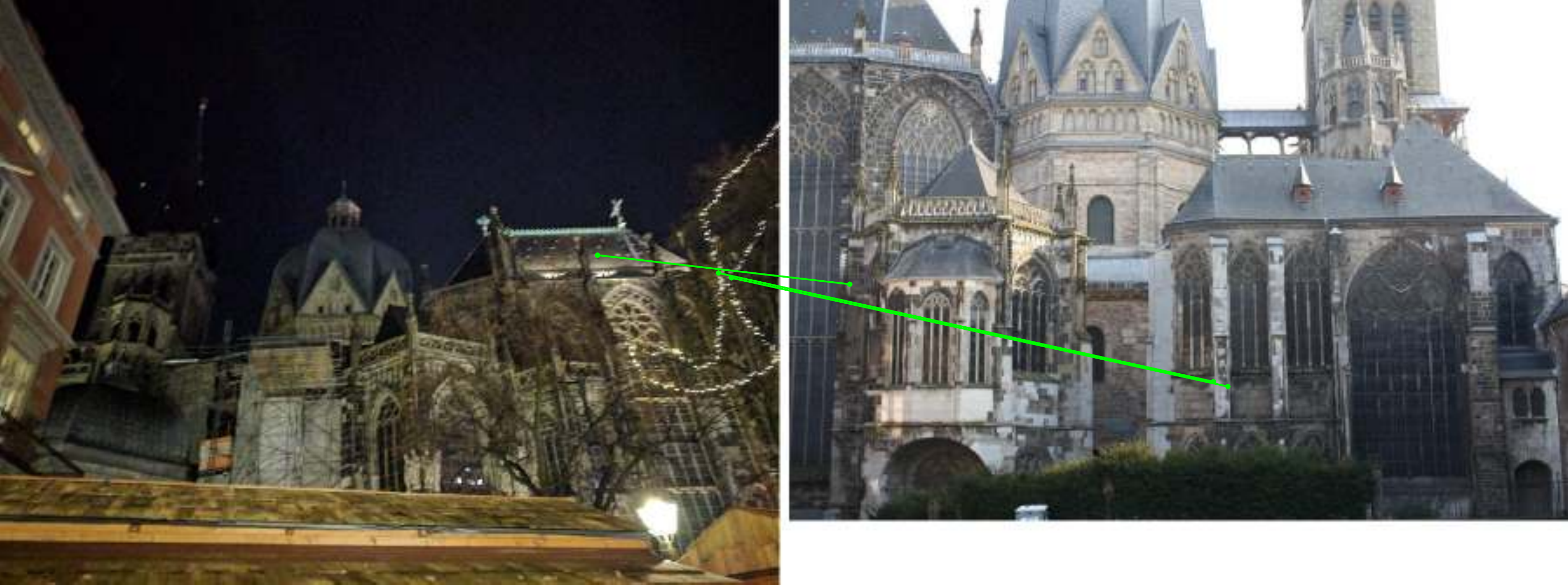}
\end{minipage}%
\vspace{3mm}

\caption{\textbf{Localization with HF-Net on Aachen night.} For each image pair, the left image is the query and the right image is the retrieved database image with the most inlier matches, as returned by PnP+RANSAC. We show challenging successful queries (left), failed queries due to an incorrect global retrieval (center), and failed queries due to incorrect or insufficient local matches (right).}
\label{fig:qual:aachen}
\end{figure*}

\begin{figure*}[htb!]
\centering
\def\iwidth{.32\linewidth}

\begin{minipage}{\iwidth}
    \centering
    \includegraphics[width=0.98\linewidth]{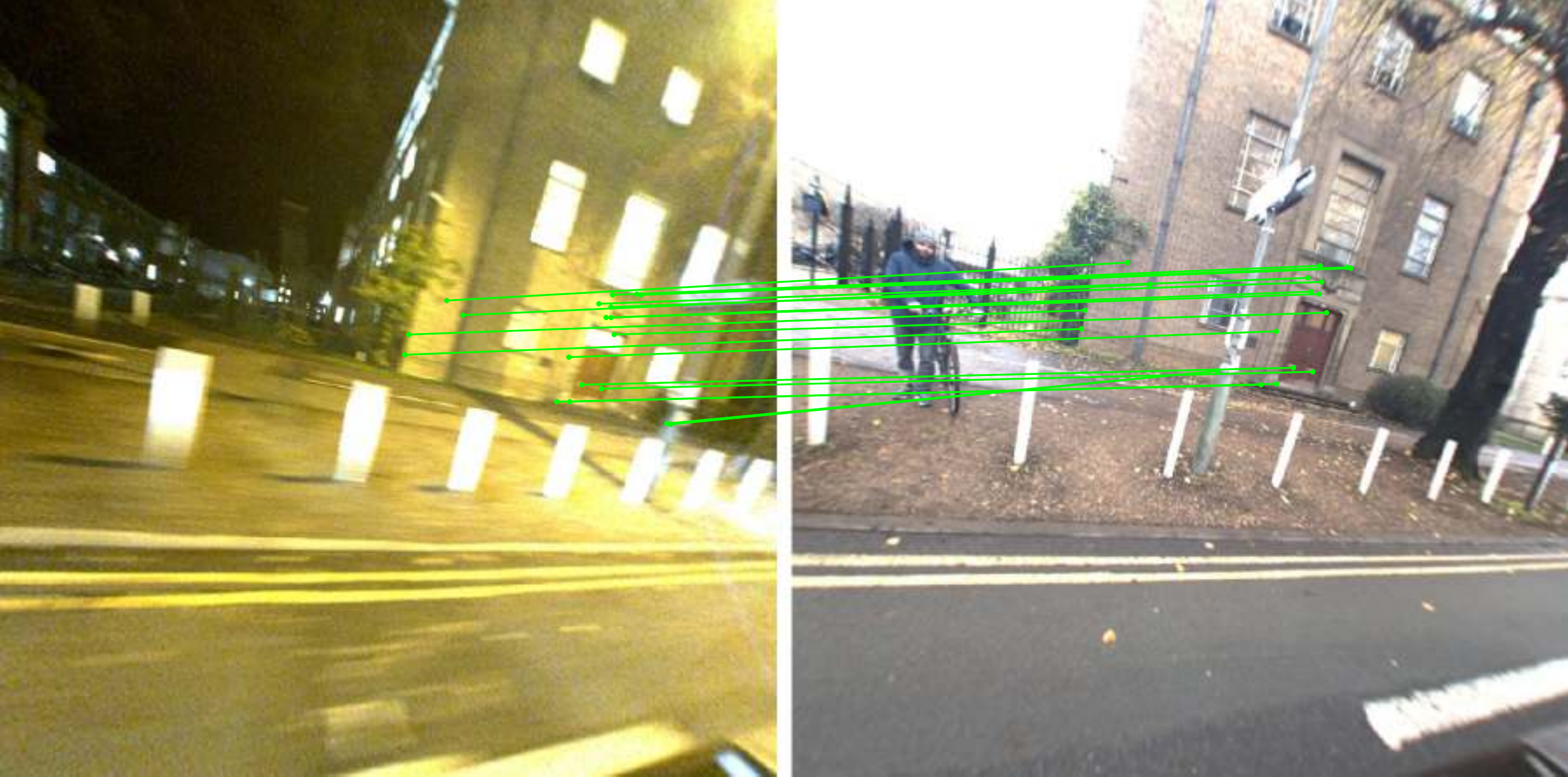}
\end{minipage}%
\begin{minipage}{\iwidth}
    \centering
    \includegraphics[width=0.98\linewidth]{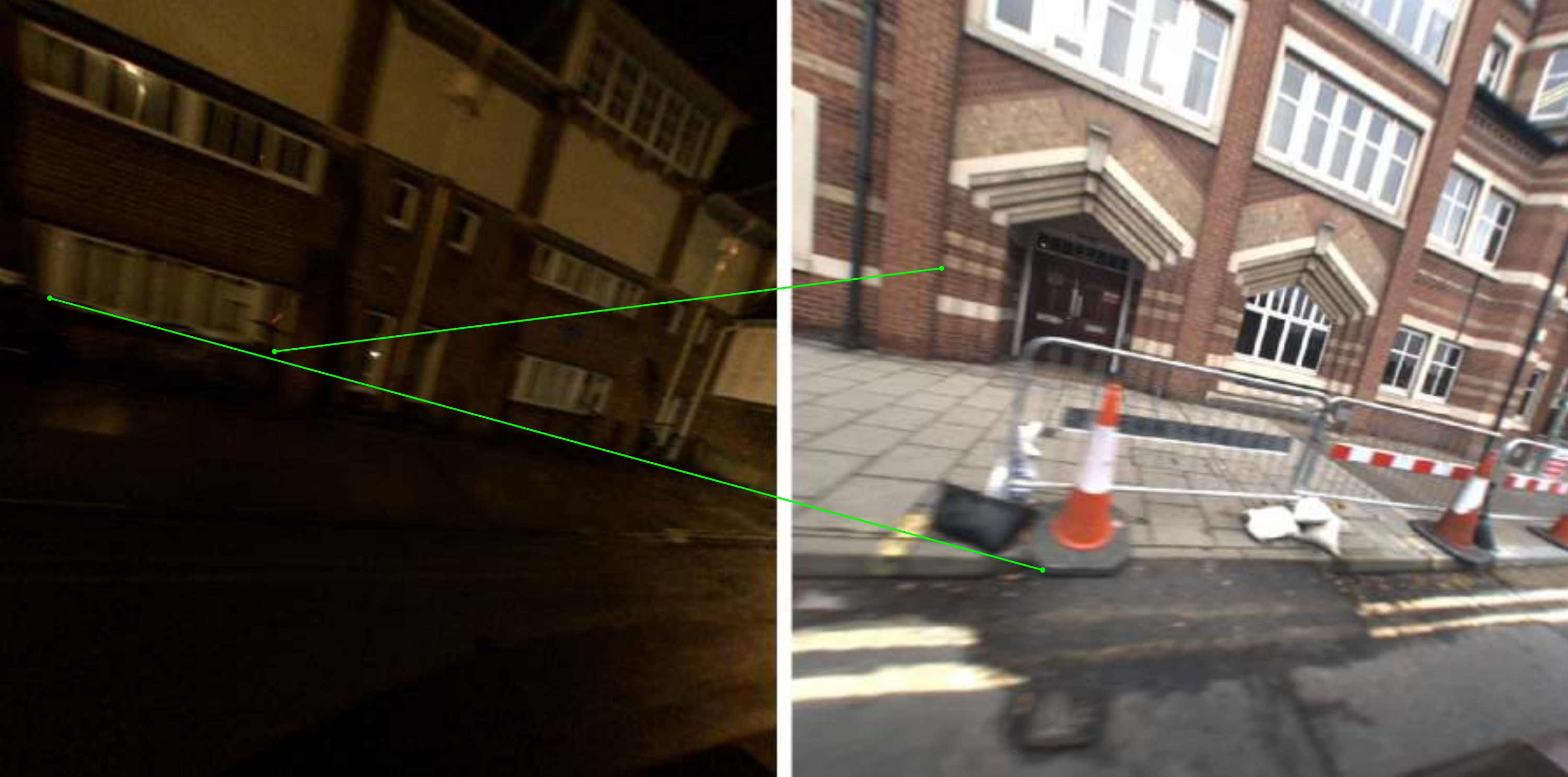}
\end{minipage}%
\begin{minipage}{\iwidth}
    \centering
    \includegraphics[width=0.98\linewidth]{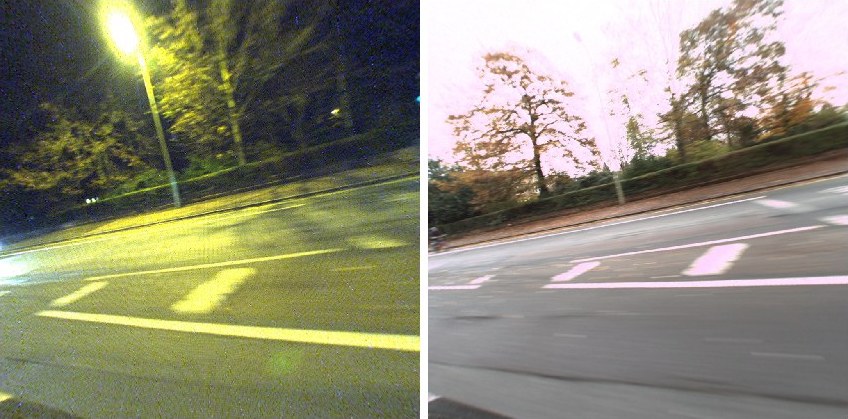}
\end{minipage}

\begin{minipage}{\iwidth}
    \centering
    \includegraphics[width=0.98\linewidth]{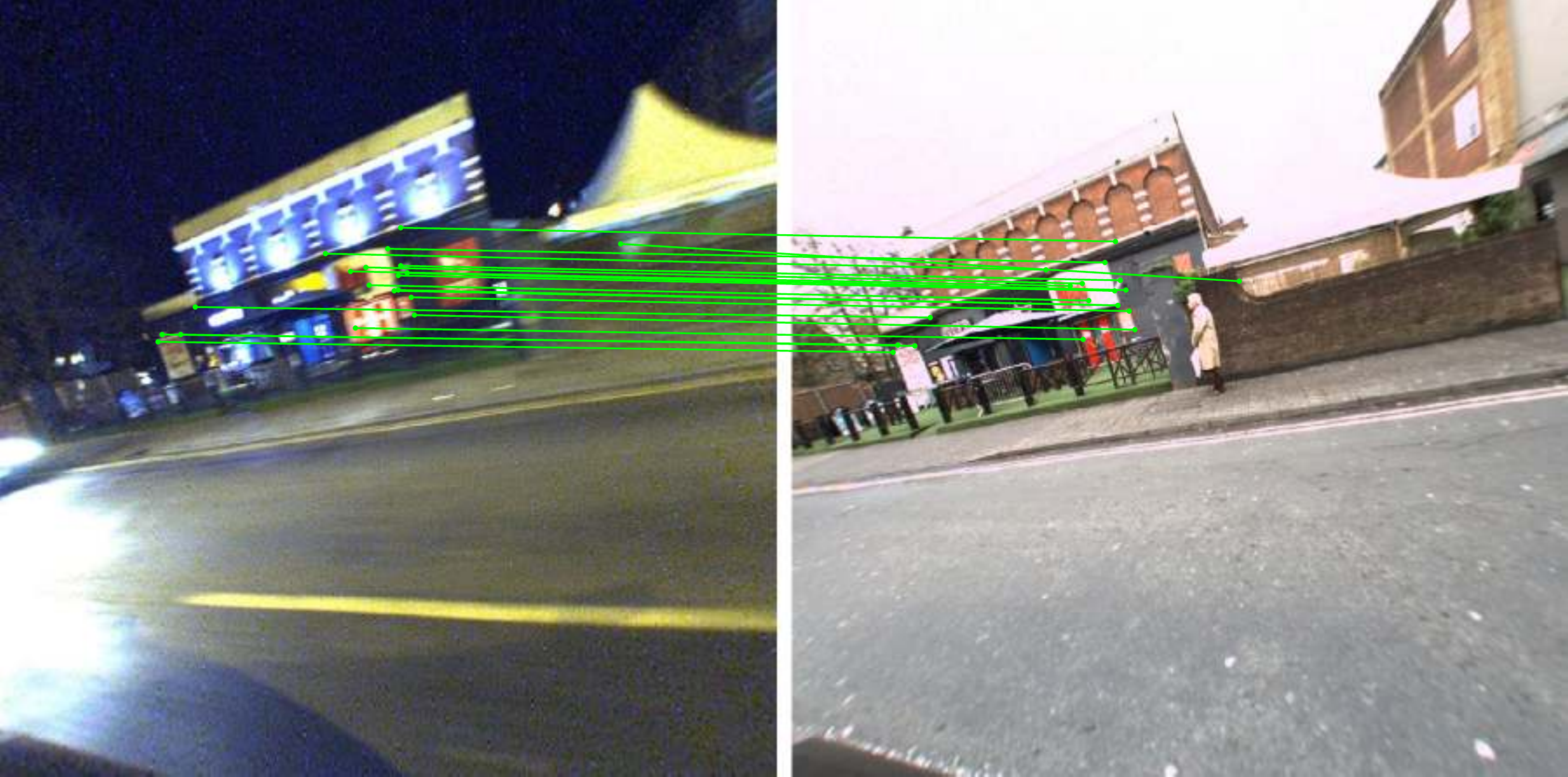}
\end{minipage}%
\begin{minipage}{\iwidth}
    \centering
    \includegraphics[width=0.98\linewidth]{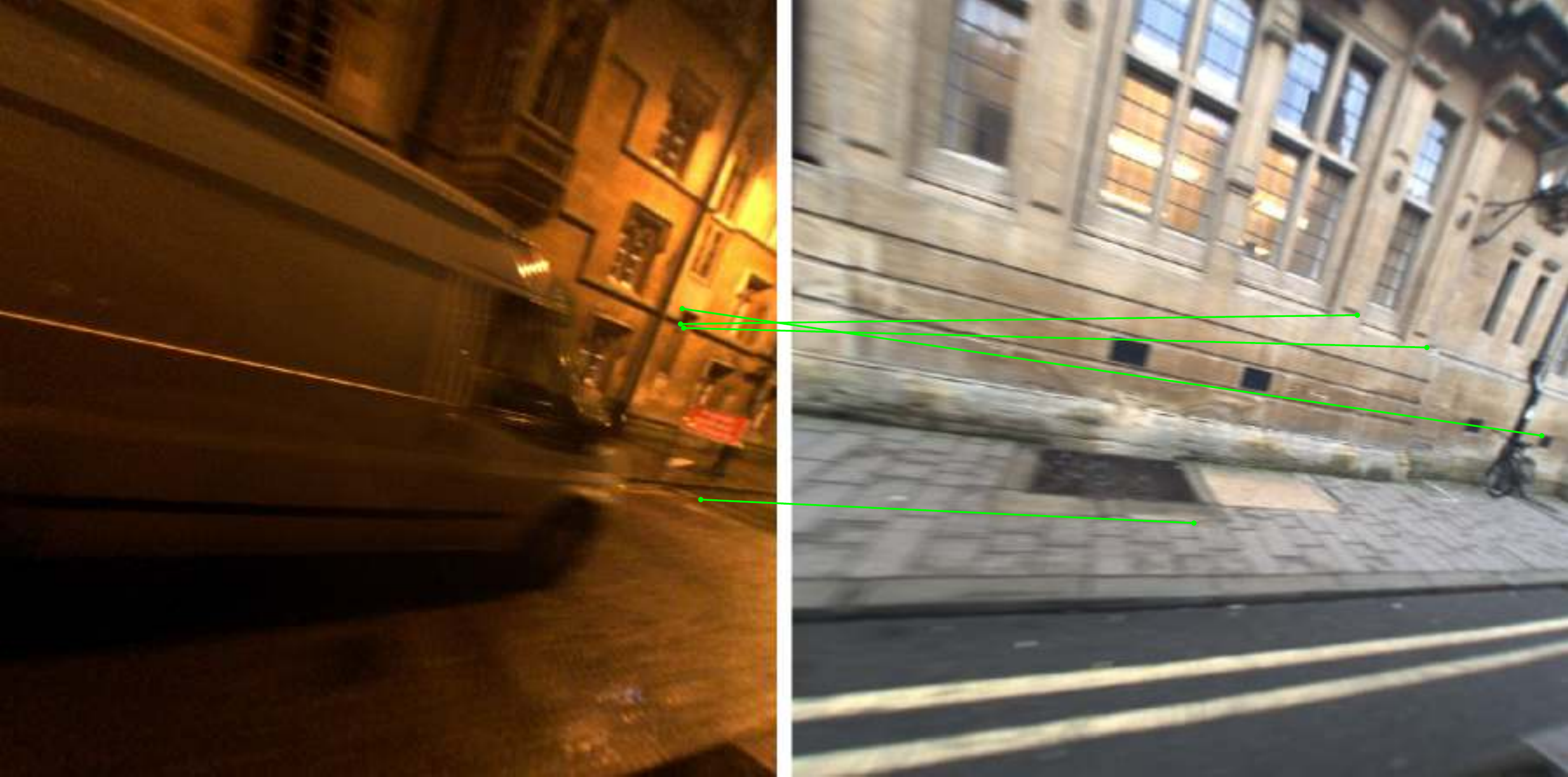}
\end{minipage}%
\begin{minipage}{\iwidth}
    \centering
    \includegraphics[width=0.98\linewidth]{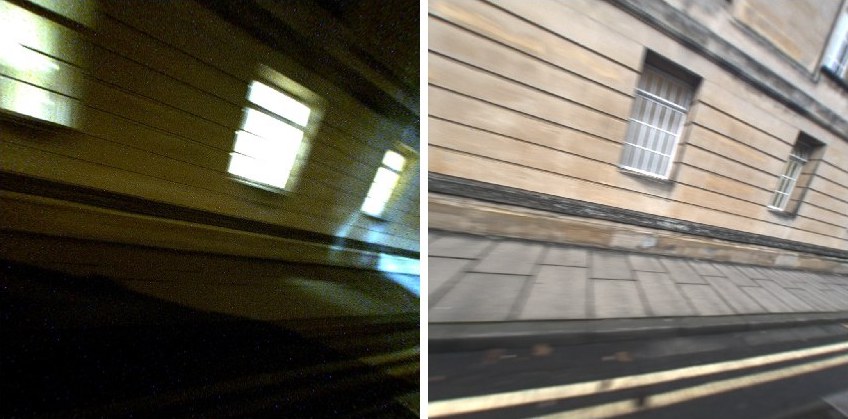}
\end{minipage}

\begin{minipage}{\iwidth}
    \centering
    \includegraphics[width=0.98\linewidth]{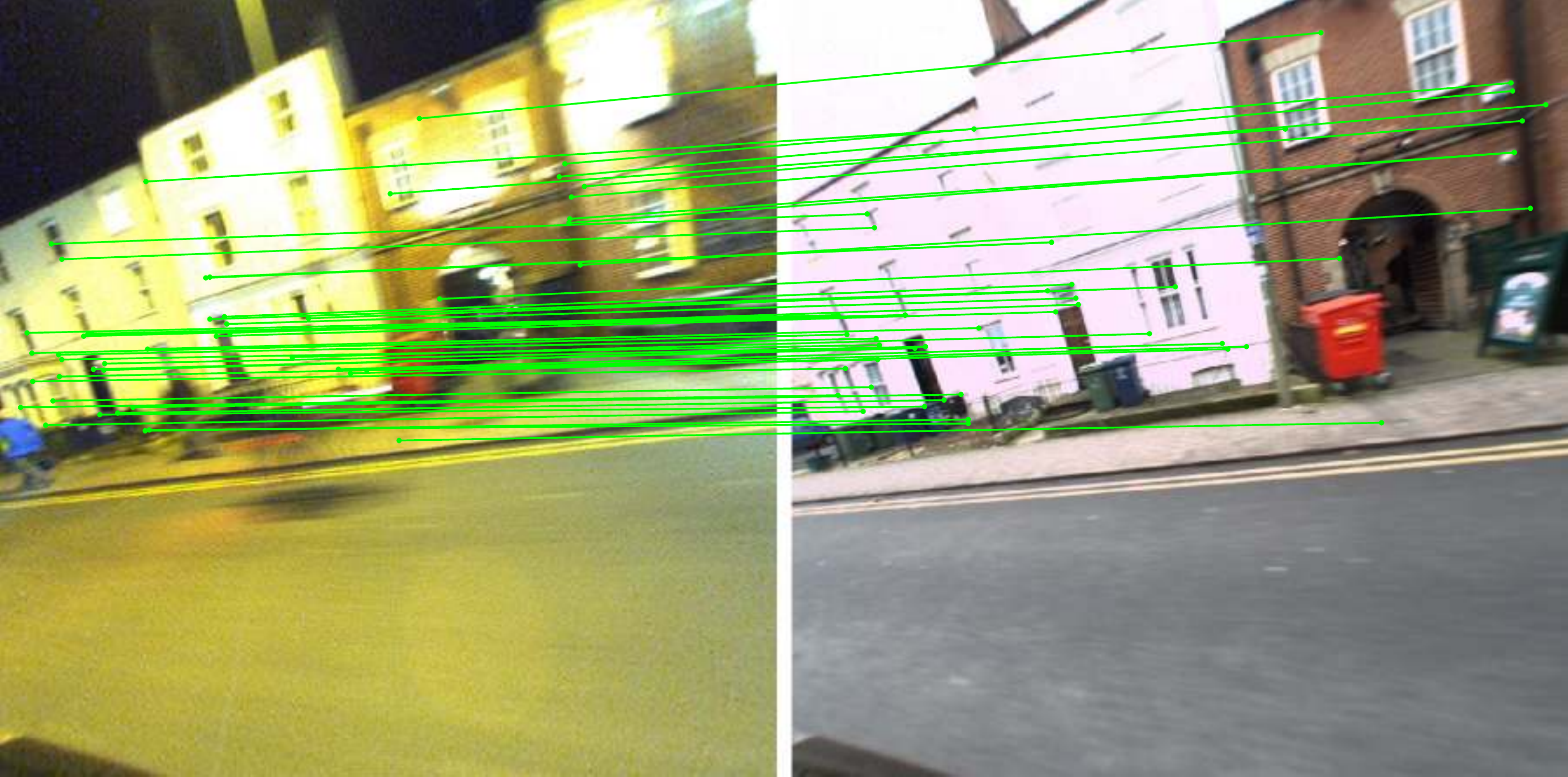}
\end{minipage}%
\begin{minipage}{\iwidth}
    \centering
    \includegraphics[width=0.98\linewidth]{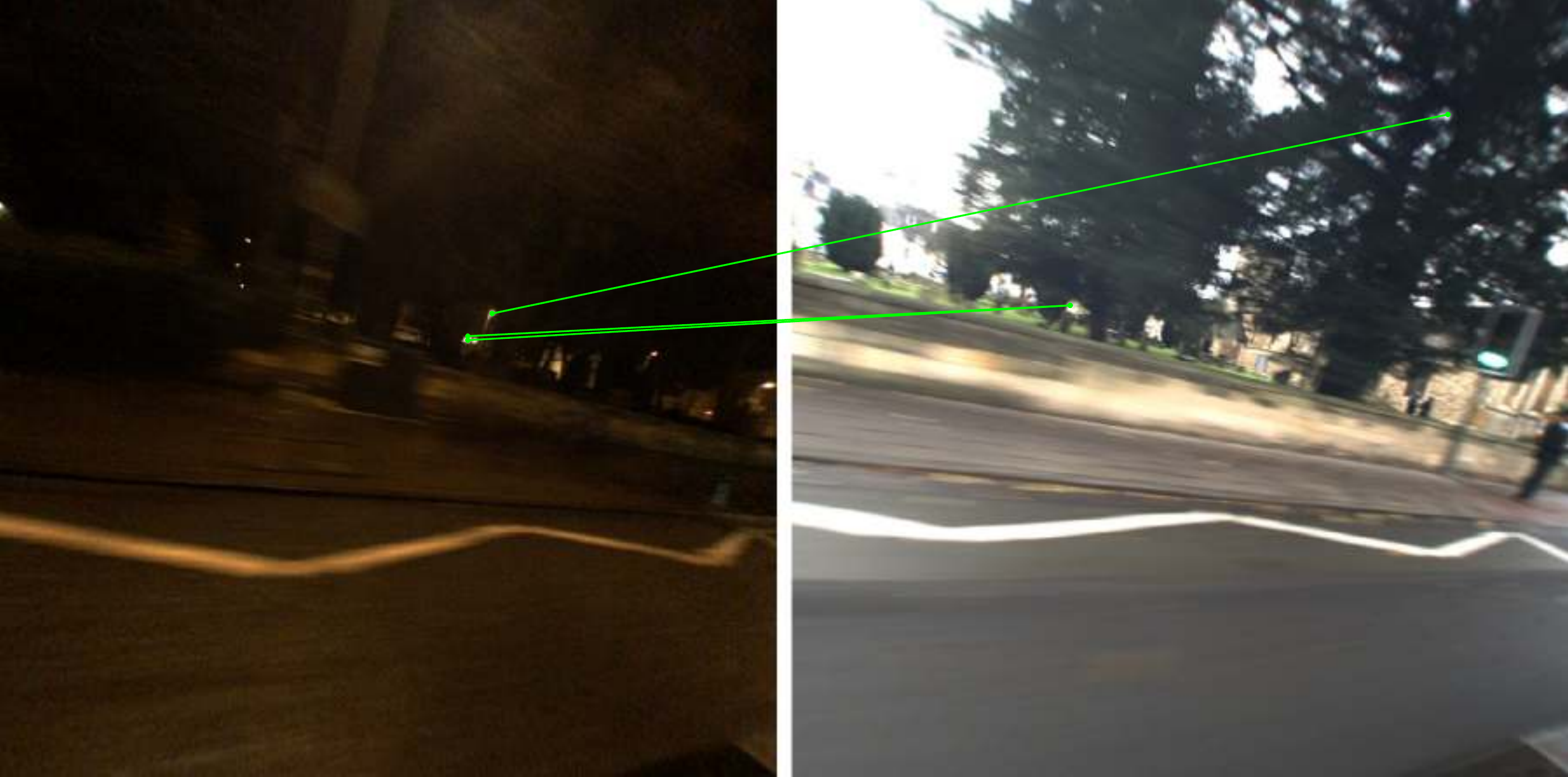}
\end{minipage}%
\begin{minipage}{\iwidth}
    \centering
    \includegraphics[width=0.98\linewidth]{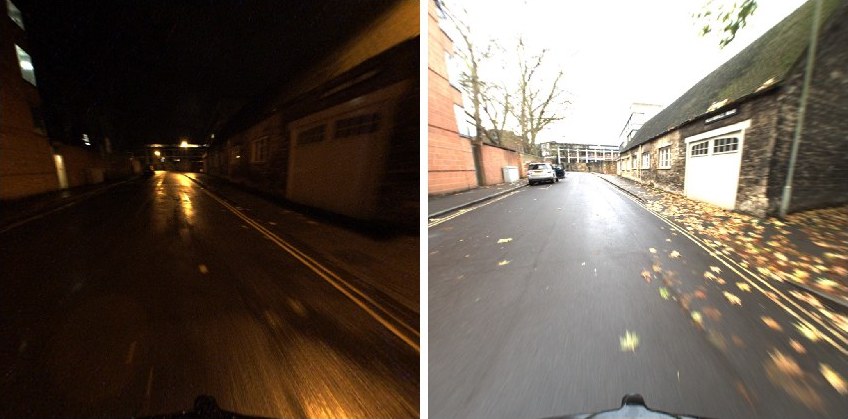}
\end{minipage}%
\vspace{3mm}

\caption{\textbf{Localization with HF-Net on RobotCar night and night-rain.} For each image pair, the left image is the query and the right image is the retrieved database image with the most inlier matches, as returned by PnP+RANSAC. We show challenging successful queries (left), failed queries due to an incorrect global retrieval (center), and failed queries due to insufficient local matches (right).}
\label{fig:qual:robotcar}
\end{figure*}

\begin{figure*}[htb!]
\centering
\def\iwidth{.32\linewidth}

\begin{minipage}{\iwidth}
    \centering
    \includegraphics[width=0.98\linewidth]{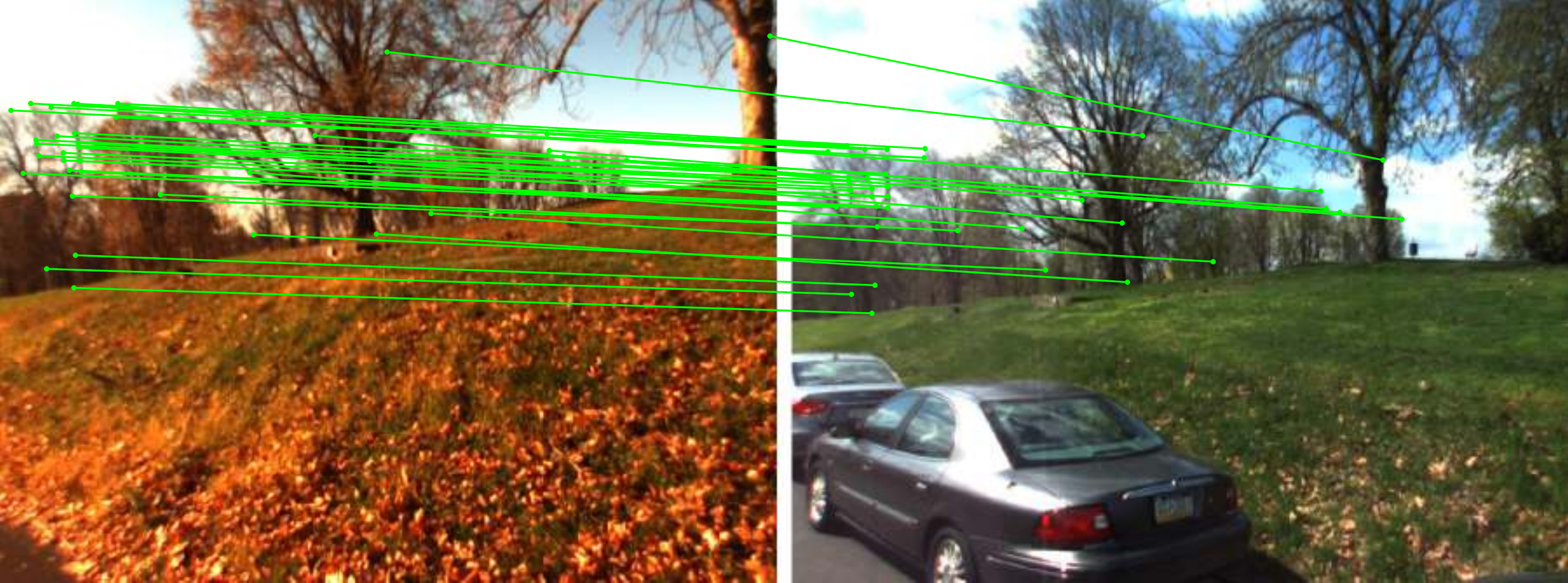}
\end{minipage}%
\begin{minipage}{\iwidth}
    \centering
    \includegraphics[width=0.98\linewidth]{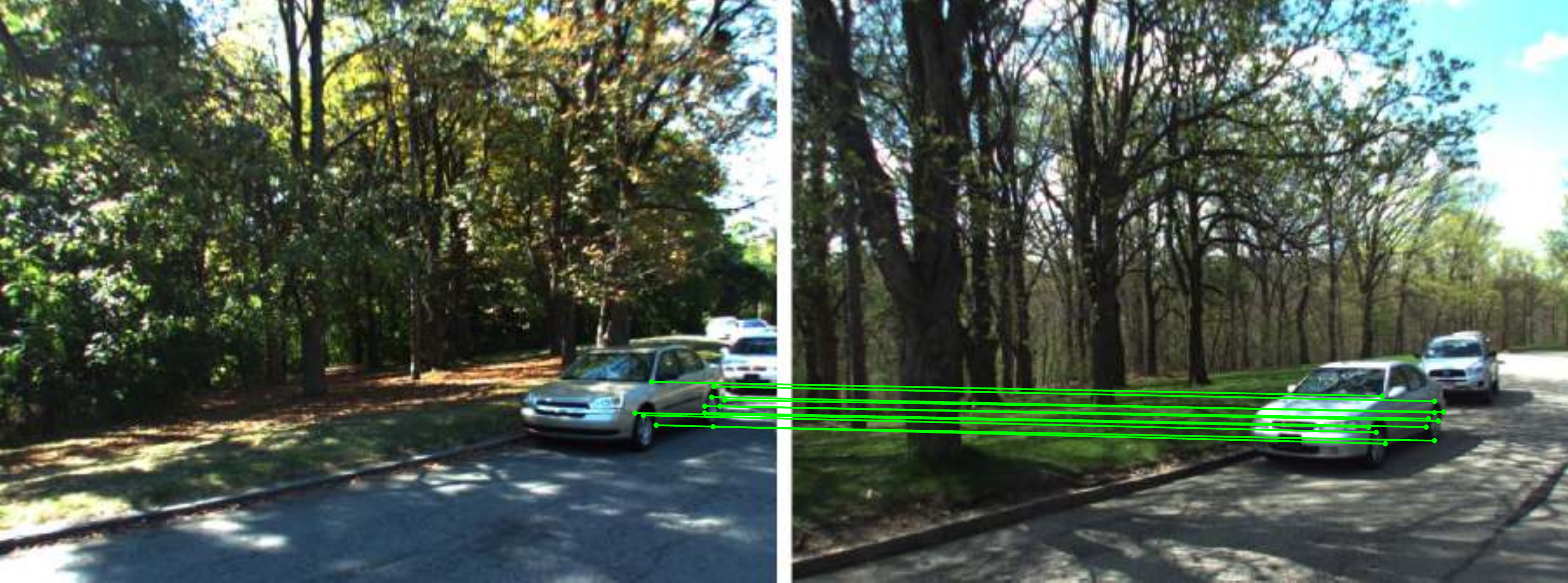}
\end{minipage}%
\begin{minipage}{\iwidth}
    \centering
    \includegraphics[width=0.98\linewidth]{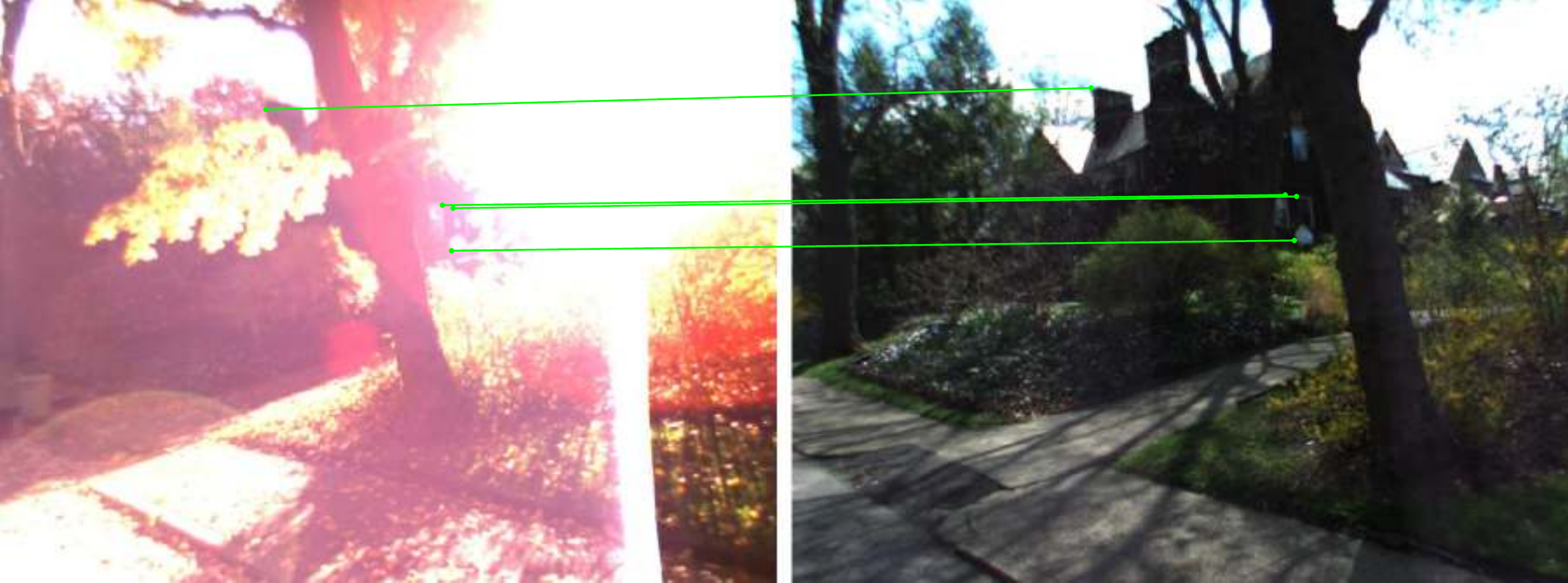}
\end{minipage}%
\vspace{2mm}

\begin{minipage}{\iwidth}
    \centering
    \includegraphics[width=0.98\linewidth]{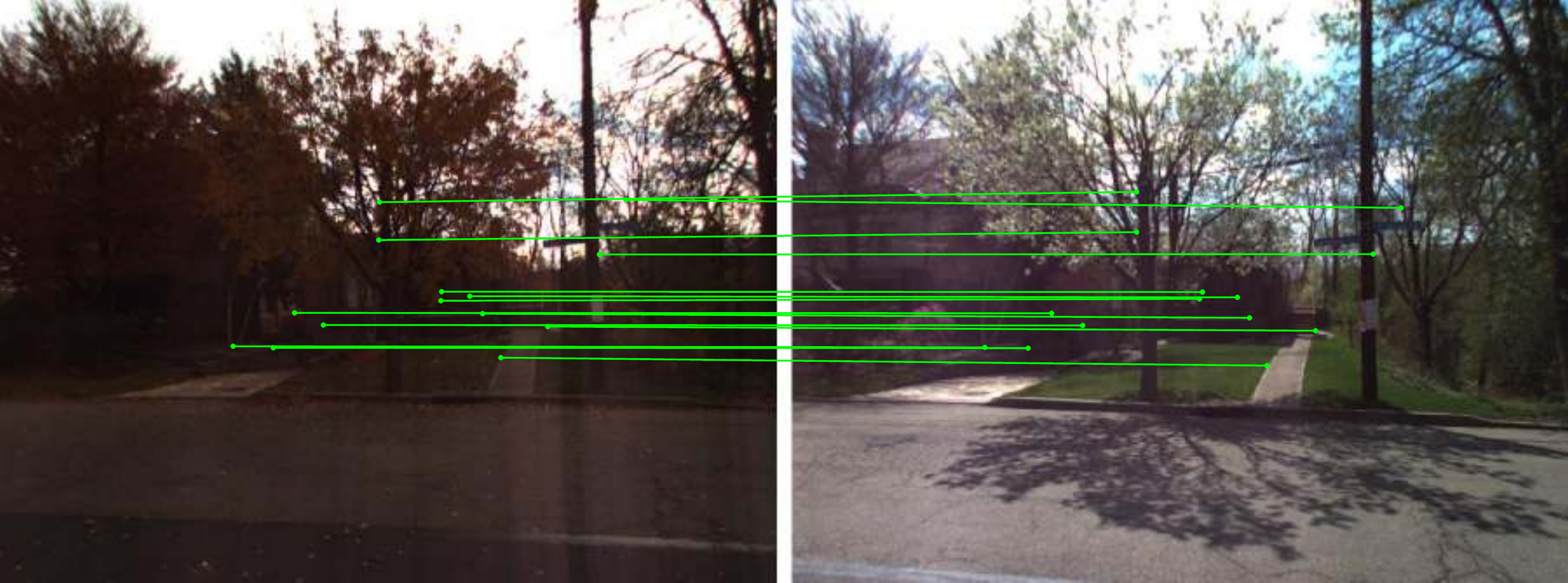}
\end{minipage}%
\begin{minipage}{\iwidth}
    \centering
    \includegraphics[width=0.98\linewidth]{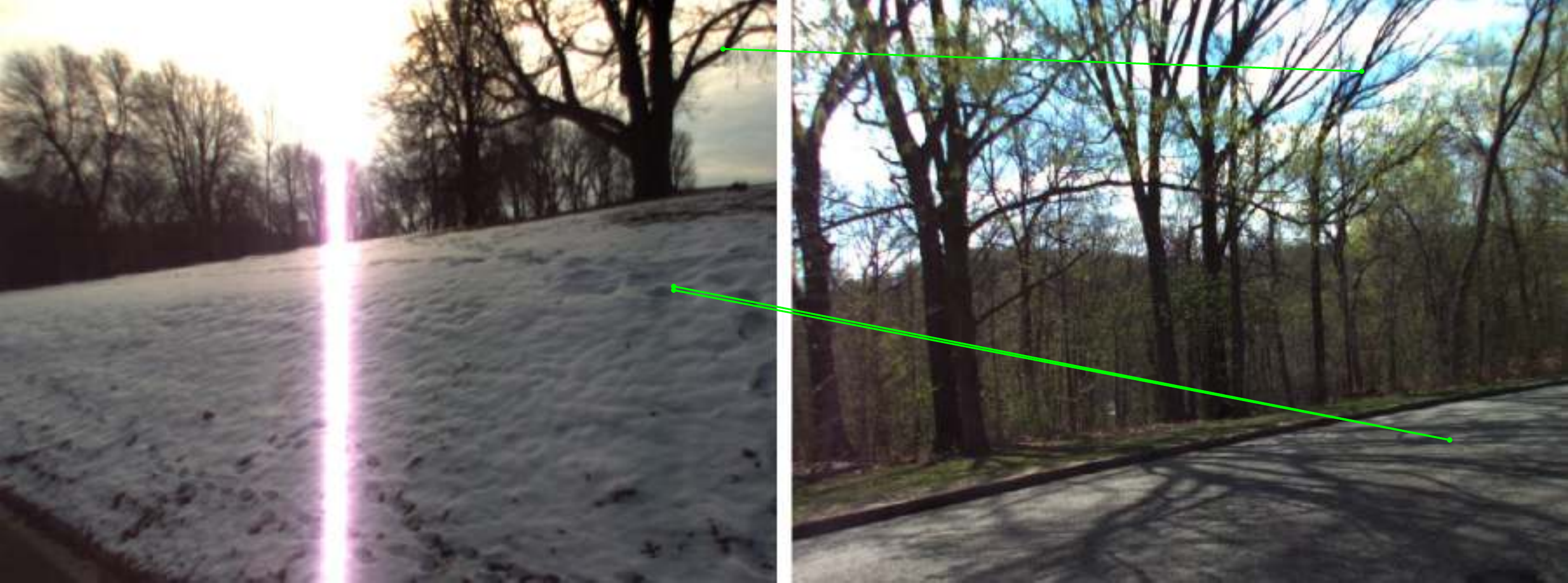}
\end{minipage}%
\begin{minipage}{\iwidth}
    \centering
    \includegraphics[width=0.98\linewidth]{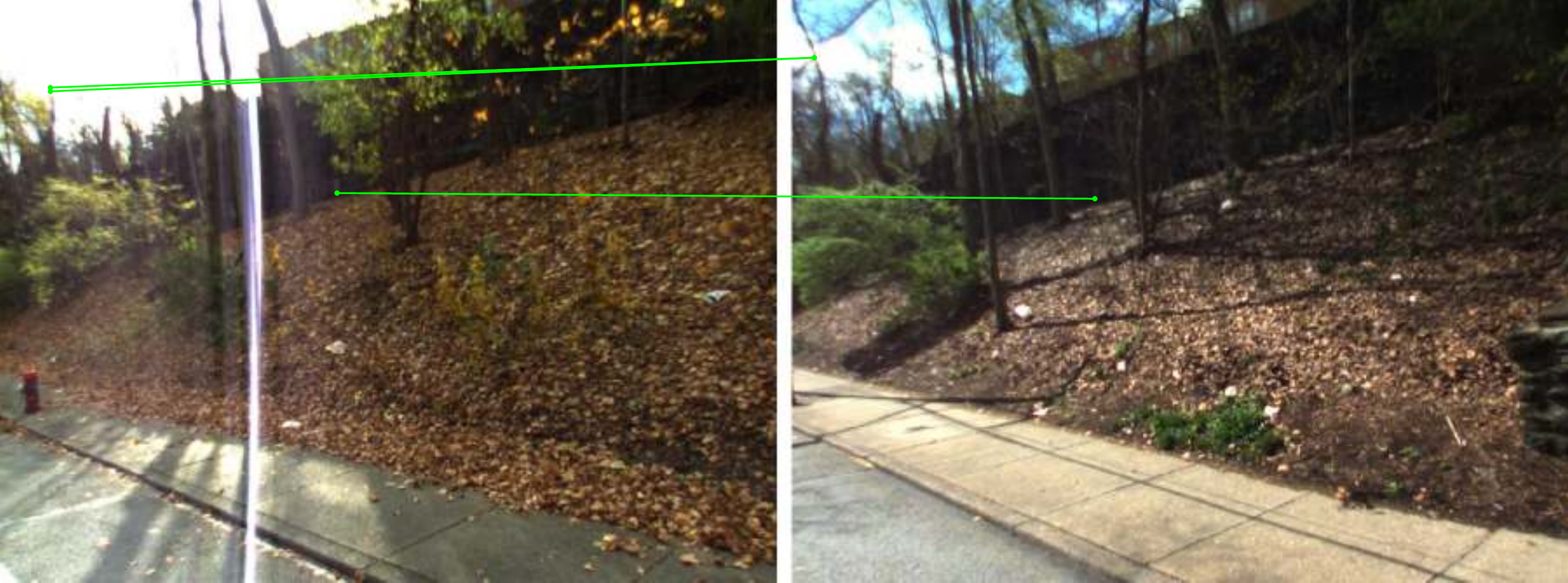}
\end{minipage}%
\vspace{2mm}

\begin{minipage}{\iwidth}
    \centering
    \includegraphics[width=0.98\linewidth]{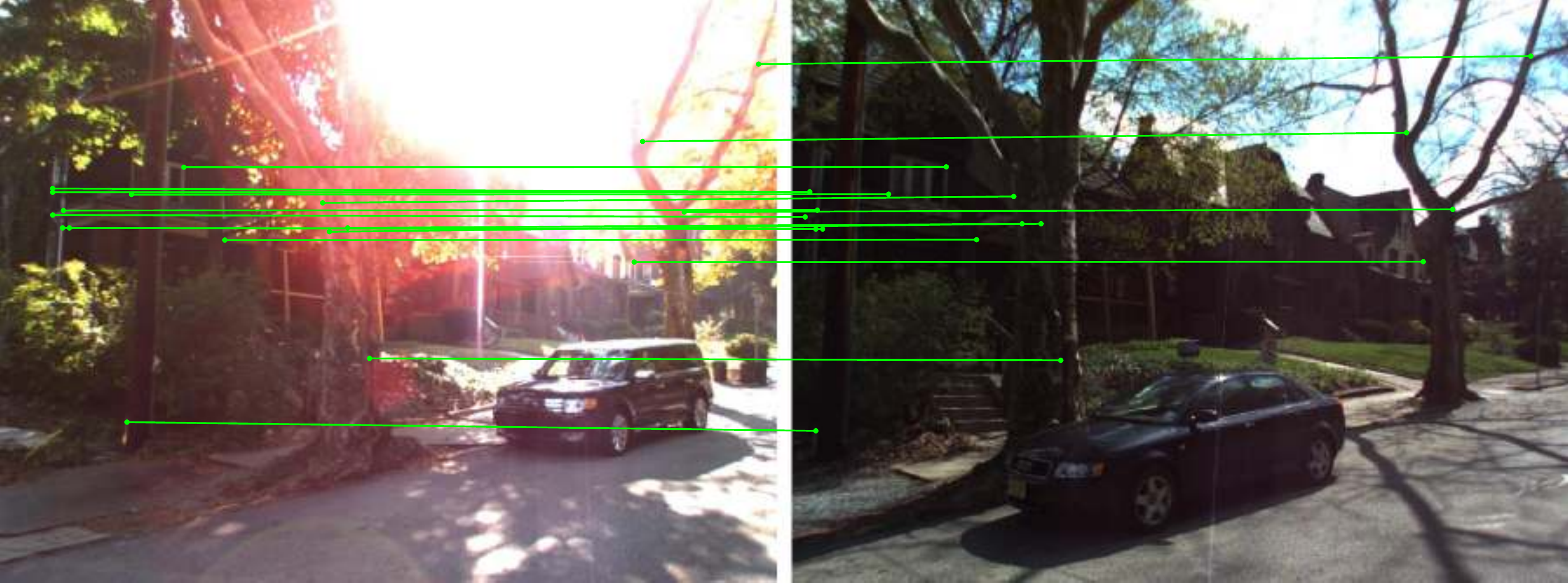}
\end{minipage}%
\begin{minipage}{\iwidth}
    \centering
    \includegraphics[width=0.98\linewidth]{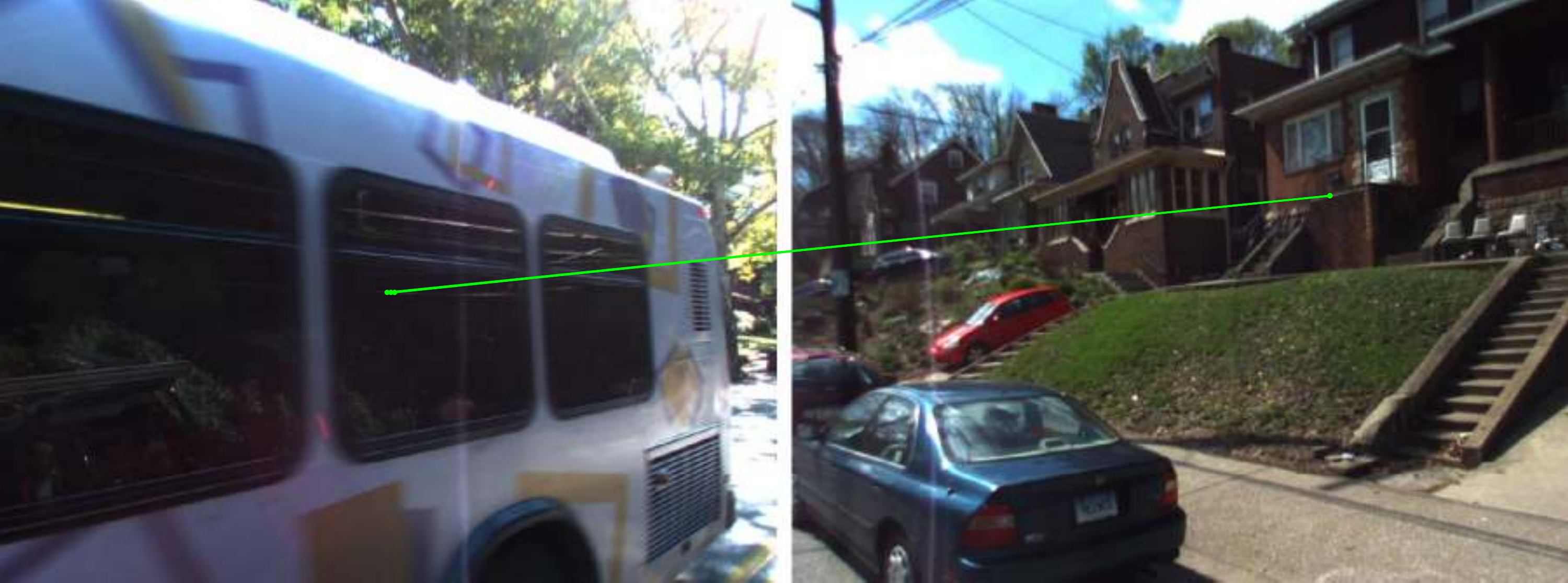}
\end{minipage}%
\begin{minipage}{\iwidth}
    \centering
    \includegraphics[width=0.98\linewidth]{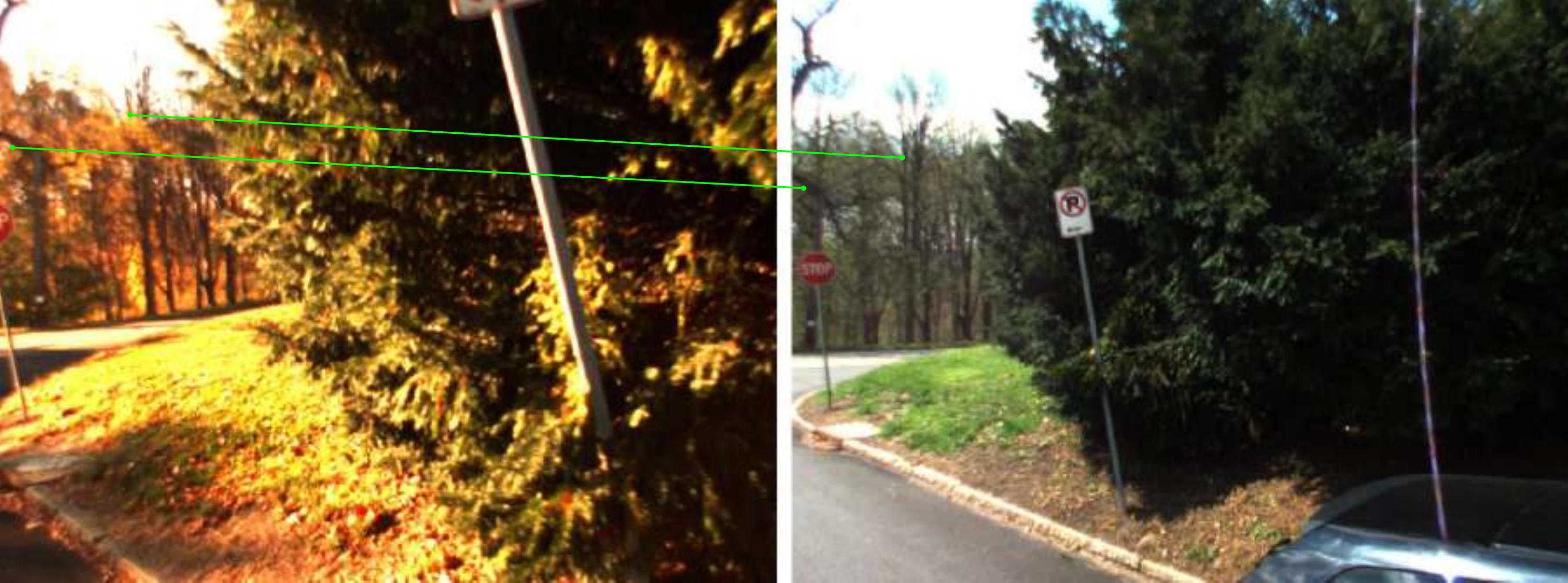}
\end{minipage}%
\vspace{3mm}

\caption{\textbf{Localization with HF-Net on CMU suburban.} For each image pair, the left image is the query and the right image is the retrieved database image with the most inlier matches, as returned by PnP+RANSAC. We show challenging successful queries (left), failed queries due to an incorrect global retrieval (center), and failed queries due to insufficient local matches (right).}
\label{fig:qual:cmu}
\end{figure*}

\begin{figure*}[htb!]
  \centering
  \def\iwidth{.30\linewidth}

  \begin{minipage}{\iwidth}
        \centering
            \includegraphics[trim=0 130 0 0,clip,width=0.98\linewidth]{images/aachen/qualitative/hfnet_night_IMG_20161227_172730_inliers.pdf}
  \end{minipage}%
  \begin{minipage}{\iwidth}
        \centering
            \includegraphics[trim=0 130 0 0,clip,width=0.98\linewidth]{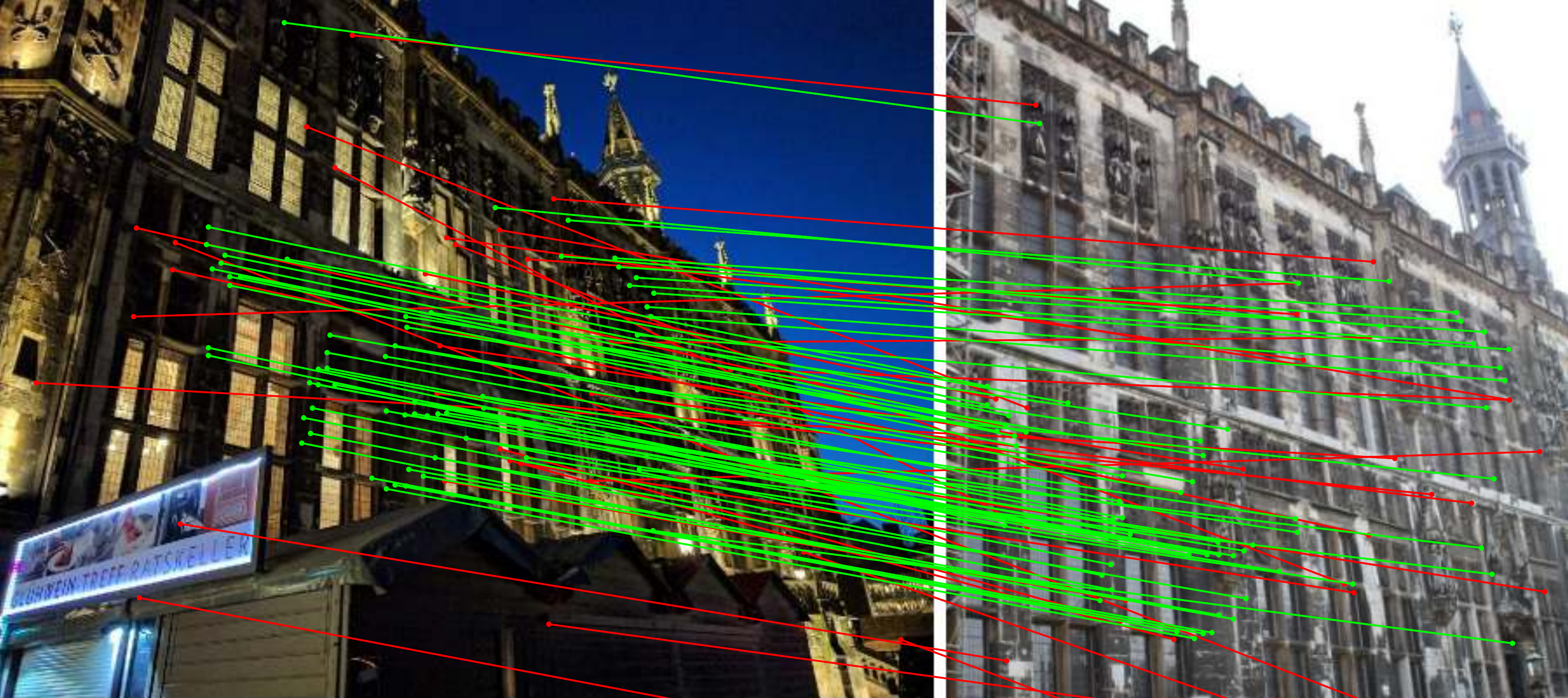}
  \end{minipage}%
  \begin{minipage}{.36\linewidth}
        \centering
            \includegraphics[width=0.98\linewidth]{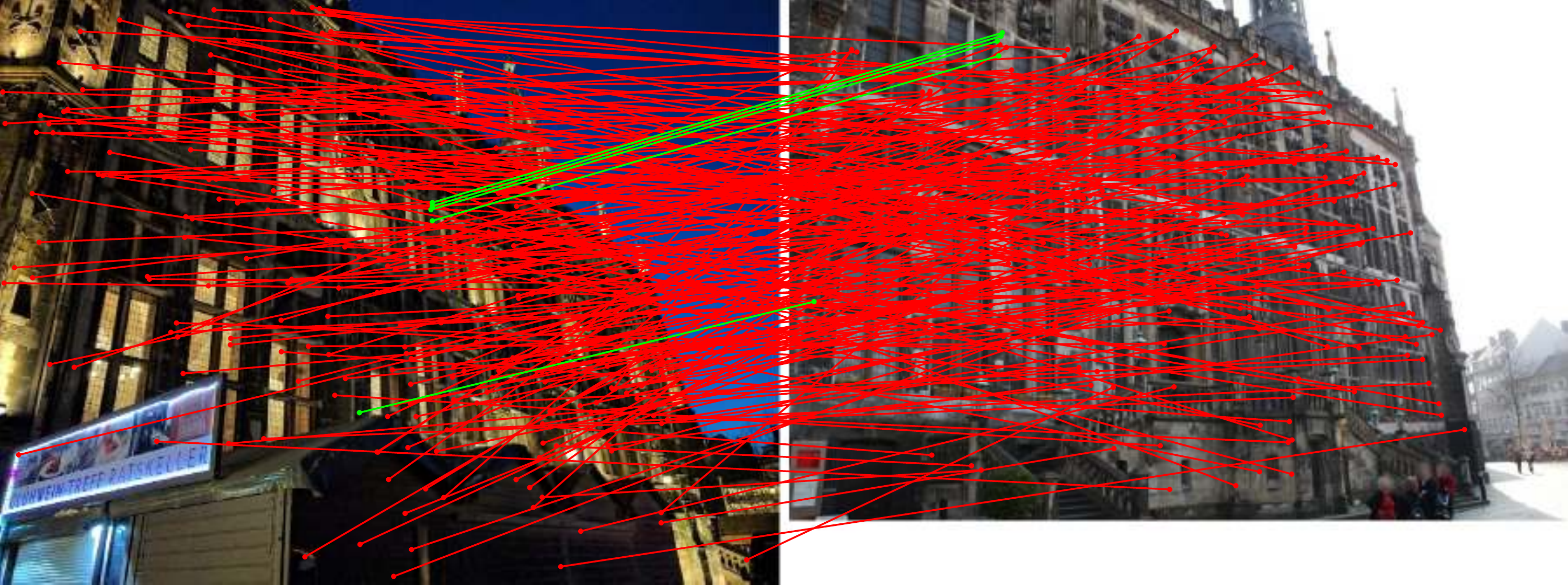}
  \end{minipage}%
  \vspace{3mm}

  \caption{\textbf{Comparison between HF-Net and NV+SIFT on Aachen night,} with one query for which HF-Net returns the correct location but NV+SIFT fails. We show the matches with one retrieved database image, labeled by PnP+RANSAC as inliers (green) and outliers (red). We show the inliers of HF-Net (left), all the matches of HF-Net (center), and all the matches of NV+SIFT (right). HF-Net generates significantly fewer matches than SIFT, thus reducing the computational footprint of the local matching.  At the same time, more of its matches are inliers, increasing the robustness of the localization. The higher inlier ratio reduces the number of required RANSAC iterations.}
  \label{fig:comp:aachen}
\end{figure*}

Visual results of HF-Net on the Aachen Day-Night, RobotCar Seasons, and CMU Seasons datasets are shown in Figures~\ref{fig:qual:aachen},~\ref{fig:qual:robotcar}, and~\ref{fig:qual:cmu}, respectively. We additionally show a comparison with NV+SIFT in Figure~\ref{fig:comp:aachen}.

{\small
\bibliographystyle{ieee_fullname}
\bibliography{ms}
}


\end{document}